\def\year{2021}\relax
\documentclass[letterpaper]{article} %
\usepackage{aaai21}  %
\usepackage{times}  %
\usepackage{helvet} %
\usepackage{courier}  %
\usepackage[hyphens]{url}  %
\usepackage{graphicx} %
\usepackage{natbib}  %
\usepackage{caption} %
\usepackage{amsmath}
\usepackage{amssymb}
\usepackage{leftidx}
\usepackage{xspace}
\makeatletter
\DeclareRobustCommand\onedot{\futurelet\@let@token\@onedot}
\def\@onedot{\ifx\@let@token.\else.\null\fi\xspace}

\def\eg{\emph{e.g}\onedot} 
\def\ie{\emph{i.e}\onedot} 
 
\def\etc{\emph{etc}\onedot}

\makeatother

\title{Consistent Right-Invariant Fixed-Lag Smoother \\
with Application to Visual Inertial SLAM}

\author {
        Jianzhu Huai \textsuperscript{\rm 1},
        Yukai Lin \textsuperscript{\rm 2},
        Yuan Zhuang
        \thanks{Corresponding author, yuan.zhuang@whu.edu.cn}\textsuperscript{, \rm 1},
        Min Shi\textsuperscript{\rm 3} \\
}
\affiliations {
    \textsuperscript{\rm 1}Wuhan University \\
    \textsuperscript{\rm 2}ETH Zurich \\
    \textsuperscript{\rm 3}Washington University, St. Louis \\
    \{jianzhu.huai,yuan.zhuang\}@whu.edu.cn, linyuk@ethz.ch,
    mshi2018@fau.edu
}

\begin{document}

\maketitle

\begin{abstract}
State estimation problems without absolute position measurements routinely arise in navigation of unmanned aerial vehicles, 
autonomous ground vehicles, \etc whose proper operation relies on 
accurate state estimates and reliable covariances.
Unaware of absolute positions, these problems have immanent unobservable directions.
Traditional causal estimators, however, usually gain spurious information 
on the unobservable directions, leading to over-confident covariance inconsistent with actual estimator errors.
The consistency problem of fixed-lag smoothers (FLSs) 
has only been attacked by the first estimate Jacobian (FEJ) technique because 
of the complexity to analyze their observability property.
But the FEJ has several drawbacks hampering its wide adoption.
To ensure the consistency of a FLS, this paper introduces the right invariant
error formulation into the FLS framework. To our knowledge, 
we are the first to analyze the observability of a FLS with the right invariant error.
Our main contributions are twofold.
As the first novelty, to bypass the complexity of analysis with the classic 
observability matrix, we show that observability analysis of FLSs can be done equivalently on the linearized system.
Second, we prove that the inconsistency issue in the traditional FLS can be
elegantly solved by the right invariant error formulation without artificially correcting Jacobians.
By applying the proposed FLS to the monocular visual inertial
simultaneous localization and mapping (SLAM) problem,
we confirm that the method consistently
estimates covariance similarly to a batch smoother in simulation and that our method achieved comparable accuracy as traditional FLSs on real data.
\end{abstract}

\section{Introduction}
Positioning and navigation of a variety of vehicles, \eg, unmanned aerial vehicles (UAVs), 
autonomous ground vehicles (AGVs), depends on real-time state estimation. 
Accurate system state and reasonable covariance output by 
state estimators in real time are necessary for the proper operation of these systems.
For state estimation, these systems usually fuse measurements 
captured by sensors that do not provide absolute positions, like cameras, lidars, inertial measurement units (IMUs), \etc.
It is well known that estimators which fuse such measurements have unobservable directions \cite{jones2011visual}.

As reported in the literature, traditional real-time estimators, \eg, filters, fixed-lag smoothers (FLSs),
tend to gain fictitious information on unobservable directions \cite{huangObservabilitybasedRulesDesigning2010,dong-siMotionTrackingFixedlag2011}, 
and to output falsely optimistic covariance inconsistent to the actual state error.
This inconsistency is caused by the marginalization step of real-time estimators
which removes old state variables and measurements (\ie, factors) from an estimator and approximates those measurements by a linear prior factor.
A deeper cause is that for a variable in the prior factor, its linearization point used by the prior factor differs from that used by the remaining factors.
Obviously, the batch estimator and its incremental variants, \eg, iSAM2 \cite{kaess2012isam2}, 
do not have this issue as they do not marginalize variables.

To fix the estimator inconsistency, techniques that modify the
measurement Jacobians to fit certain criteria have been proposed. 
For instance, the ``first estimate Jacobian (FEJ)'' technique \cite{huangObservabilitybasedRulesDesigning2010} 
evaluates Jacobians relative to variables in the linear prior factor at their estimates upon marginalization.
Because the Jacobian computation depends on specifics, such as an earlier estimate of a variable,
it is usually difficult to apply such techniques to an existing estimator framework.
A new trend is to use right invariant error formulation \cite{barrau_ekf_2016} where
a navigation state variable (consisting of orientation, position, and velocity) is associated to a Lie group $SE_2(3)$ 
and the error vector is invariant to transforming the trajectory by a right multiplication.
Besides mathematically elegant, it is easy to implement as it fits the conventional filtering framework.
However, this formulation has not been used in FLSs, mainly because of the challenge to analyze their consistency property.

Previous work has shown that the estimator inconsistency comes along with the observability issue where the unobservable directions become spuriously observable \cite{heschCameraIMUbasedLocalizationObservability2014}.
Thus, consistency has been predominantly studied by examining rank deficiency of the linearized observability matrix, \eg,
\cite{huangObservabilitybasedRulesDesigning2010,dong-siConsistencyAnalysisSlidingwindow2012,brossardExploitingSymmetriesDesign2018}.
The local observability matrix is acceptable in complexity for analyzing filters, 
but becomes very involved for dealing with FLSs, \eg, \cite{dong-siConsistencyAnalysisSlidingwindow2012}.
Because the observability matrix is a derivative of the linearized original system, 
we think that directly working with the linearized system can greatly simplify the observability analysis.

Based on this analysis, we prove that the right invariant error formulation leads to a consistent FLS.
The claims made in the proof are validated with simulation.
Furthermore, the practicality of the proposed right invariant FLS is verified with the EuRoC benchmark \cite{burri2016euroc}.

In summary, our contributions include
\begin{itemize}
	\item To avoid the complexity of observability matrices, 
	we prove that observability analysis of FLSs can be done equivalently on the linearized system.
	\item To clarify effects of variables on observability, we show that using
	different linearization points for a state variable expressed in a local coordinate frame 
	and for a sensor parameter do not impact unobservable directions and hence consistency.
	\item To our knowledge, we are the first to prove and validate
	that FLS with the right invariant error formulation maintains consistent covariance without artificially modifying Jacobians.
\end{itemize}

The following text presents the formulation and observability analysis of the FLS to solve the visual inertial SLAM problem
and the application of right invariant errors in the FLS.
Then, results of simulation and real data tests are supplied.
Lastly, we draw conclusions and indicate future work.

\section{Related Work}
There are several approaches to ensure consistency of traditional real-time estimators.
Most of them are designed for Extended Kalman Filters (EKFs) and 
few are proposed for optimization-based approaches, \ie, FLSs.
\citeauthor{costanteUncertaintyEstimationDatadriven2020} \citeyear{costanteUncertaintyEstimationDatadriven2020} developed a deep neural network to output state estimates and uncertainty measures, 
but their consistency is very challenging to analyze.
The optimization-based iSAM2 \cite{kaess2012isam2} method updates only affected variables as new
observations arrive, keeping constant computation cost. 
As it keeps the entire history of variables and observations for inference, 
its consistency naturally follows.
But it will drain the memory in a long-term operation.
For EKFs, the consistency remedies include robocentric coordination \cite{castellanosRobocentricMapJoining2007}, 
FEJ \cite{li_high_2013}, observability constraints \cite{heschConsistencyAnalysisImprovement2014},
and the recently developed right invariant error formulation \cite{barrau_ekf_2016,zhangConvergenceConsistencyAnalysis2017,
heoConsistentEKFbasedVisualinertial2018}.
The invariant error formulation for filters defines the error state in an extended Lie group 
such that the error state is independent of the state variable's linearization point.
As a result, the inconsistency caused by using different linearization points for the same state variable is prevented.

For FLSs, to our knowledge, 
their consistency has only been improved with the FEJ technique \cite{dong-siMotionTrackingFixedlag2011}.
But the FEJ technique for FLSs has several downsides.
The obvious one is that Jacobian matrices required by the estimators are evaluated
at less accurate earlier estimates of state variables which may adversely affect state estimation accuracy.
Second, it is often confusing to tell which state variable should lock its linearization point, 
and which Jacobian should be computed with these preset linearization points.
For instance, \citeauthor{li_high_2013} \citeyear{li_high_2013} locked linearization points for only position and velocity, 
and \citeauthor{usenko_visual_2020} \citeyear{usenko_visual_2020} locked linearization points for position, velocity,
and biases once they are in the prior factor.
Third, assigning and tracking linearization points requested by FEJ is often
impossible for generic nonlinear solvers without hacking. 
For example, a solver may encapsulate state variables such that they are not tampered by external assignment.
Recently, the left invariant error formulation has been used in a FLS
\cite{brossardAssociatingUncertaintyExtended2020} but for the purpose of uncertainty propagation on the extended Lie group $SE_2(3)$.

To analyze the observability of an estimator, there are in general two categories of approaches:
those based on the linearized observability matrix of the discrete system, 
and those based on the observability matrix built from Lie derivatives of the continuous-time system.
The discrete analysis is suitable to identify unobservable directions under a degenerate motion,
and the Lie differentiation analysis is suitable to identify the requirements to make all state variables observable.
Other methods exist but are typically unsuitable to examine the interplay between observability and consistency \cite{heschCameraIMUbasedLocalizationObservability2014}.
The first category includes 
\cite{huangObservabilitybasedRulesDesigning2010,li_high_2013,heschConsistencyAnalysisImprovement2014,dong-siConsistencyAnalysisSlidingwindow2012,zhangConvergenceConsistencyAnalysis2017,yangOnlineImuIntrinsic2020}.
The second category includes \cite{mirzaei_kalman_2008,kelly2011visual,heschCameraIMUbasedLocalizationObservability2014,jungObservability2020}.
The conclusions by methods from the two categories are congruent.
Interestingly, in examining observability, all cited methods parameterize landmarks in the world frame 
rather than in a local camera frame, possibly to reduce complexity.
However, we find that expressing landmarks in a local frame is actually advantageous to the consistency analysis.

\section{Methodology}
\label{sec:method}
This section presents the proposed right invariant FLS applied to the visual inertial SLAM problem with the analysis of its consistency.
Though many state estimation problems without absolute position measurements may exhibit the inconsistent issue in an estimator,
\eg, stereo visual odometry \cite{dong-siConsistencyAnalysisSlidingwindow2012},
we choose to analyze the visual inertial SLAM problem regarding consistency 
because its observability property has been well studied and well-known.
Though some variables in the following discussion are specific to the visual inertial
SLAM, the proposed method for ensuring consistency is generic enough to translate to other state estimation problems solvable by a FLS.

We first formulate the visual inertial SLAM problem from the perspective of a FLS.
Second, we present the right invariant error formulation, and prove that the consistency property is guaranteed.

\subsection{Visual Inertial SLAM Formulation}
In a typical visual inertial SLAM problem, we try to estimate the platform state,
sensor parameters, and the unknown positions of landmarks in the environment, 
by fusing data captured by at least one camera and an IMU rigidly mounted on the platform.

\subsubsection{State Variables}
The state of the system at time $t_i$ consists of the navigation state of the platform $\boldsymbol{\pi}_i$ and the IMU biases $\mathbf{b}_i$, \ie, $\mathbf{x}_i = (\boldsymbol{\pi}_i, \mathbf{b}_i)$.
In turn, a navigation state $\boldsymbol{\pi}_i$ includes orientation $\mathbf{R}_i$,
velocity $\mathbf{v}_i$, and position $\mathbf{p}_i$ of the body frame
\{$B$\} (affixed to the platform) expressed in a world frame \{$W$\} ($z$-axis along gravity),
\ie, $\boldsymbol{\pi}_i = (\mathbf{R}_i, \mathbf{v}_i, \mathbf{p}_i)$.
For clarity, the considered sensor parameters are only the IMU biases
$\mathbf{b}$ which includes the gyro bias $\mathbf{b}_g$ and the accelerometer bias $\mathbf{b}_a$, \ie, $\mathbf{b} = (\mathbf{b}_g, \mathbf{b}_a)$.
We denote by $\mathbf{x}_{0:k}$ the entire history of system states up to time $t_k$, \ie, $\mathbf{x}_{0:k} = \{\mathbf{x}_i | i = 0, 1, \dots, k\}$.

The SLAM problem also estimates landmark feature positions $\mathbf{f}_l$, each of which is represented by an inverse depth
parameterization in an anchor camera frame \{$C_a$\} \cite{civeraInverseDepthParametrization2008}, \ie, 
\begin{equation}
\mathbf{f}_l = [\alpha, \beta, 1, \rho]^\intercal = [x/z, y/z, 1, 1/z]^\intercal
\end{equation}
where $[x, y, z]^\intercal$ is the Cartesian coordinates of the landmark in \{$C_a$\}.
The inverse depth parameterization is chosen for two reasons. 
First, it has been shown to outperform the traditional Euclidean parameterization \cite{sola2012impact,polokEffect2015}.
Second, it decouples the landmark parameters from the platform pose in the world frame,
thus they remain invariant under Euclidean transform of the original problem and
have no bearing on the observability analysis.

We denote by $\mathcal{X}_k$ the history of state variables up to $t_k$,
\begin{equation}
\mathcal{X}_{k} = \{\mathbf{x}_i | i = 0, 1, \dots, k\} \cup \{\mathbf{f}_l | l = 1, 2, \dots, L\}.
\end{equation}

\subsubsection{Measurements}
Measurements in the visual inertial SLAM problem include camera observations and IMU readings.
The observation $\mathbf{z}_{il}$ of a landmark $\mathbf{f}_l$ in camera frame \{$C_i$\} at $t_i$ is
represented by a projection model $\mathbf{h}$ which encodes the camera intrinsic parameters, \ie,
\begin{equation}
\label{eq:projection}
	\mathbf{z}_{il} = \mathbf{h}(\mathbf{T}_{BC}^{-1}\mathbf{T}_{WBi}^{-1} \mathbf{T}_{WBa} \mathbf{T}_{BC} \mathbf{f}_l) + \mathbf{n}_c
\end{equation}
where $\mathbf{n}_c \sim N(0, \boldsymbol{\Sigma}_c)$ is 2D Gaussian noise of covariance $\boldsymbol{\Sigma}_c$,
$\mathbf{T}_{BC} \in SE(3)$ is the camera extrinsic parameters, 
and $\mathbf{T}_{WBi} = (\mathbf{R}_{WBi}, \mathbf{p}_{WBi}) = (\mathbf{R}_{i}, \mathbf{p}_{i})$ and $\mathbf{T}_{WBa}$ are the platform poses 
at the observing epoch $t_i$ and the anchor epoch $t_a$.
Without loss of generality, we assume $\mathbf{T}_{BC}$ is well
calibrated and known. 
Considering that $\mathbf{T}_{WBi}$ and $\mathbf{T}_{WBa}$ are subsumed by
$\mathbf{x}_i$ and $\mathbf{x}_a$, the projection model can also be written as 
$\mathbf{z}_{il} = \mathbf{h}(\mathbf{x}_i, \mathbf{x}_a, \mathbf{f}_l) + \mathbf{n}_c$.

In a simplified IMU model, the IMU measurements $\mathbf{a}_m$ and
$\boldsymbol{\omega}_m$ are assumed to be affected by accelerometer and gyroscope biases,
$\mathbf{b}_a$ and $\mathbf{b}_g$, and Gaussian white noise processes,
$\boldsymbol{\nu}_{a}$ and $\boldsymbol{\nu}_{g}$, of power spectral densities,
$\sigma^2_a\mathbf{I}_3$ and $\sigma^2_g \mathbf{I}_3$, respectively, \ie,
\begin{align}
\label{eq:simple_accel_model}
\mathbf{a}_m &= \leftidx_{B}\mathbf{a}_s + \mathbf{b}_a + \boldsymbol{\nu}_a \\
\dot{\mathbf{b}}_a &= \boldsymbol{\nu}_{ba} \\
\label{eq:simple_gyro_model}
\boldsymbol{\omega}_m &= \leftidx_{B}\boldsymbol{\omega}_{WB} +
\mathbf{b}_g + \boldsymbol{\nu}_g \\
\dot{\mathbf{b}}_g &= \boldsymbol{\nu}_{bg},
\end{align}
where the biases are assumed to be driven by Gaussian white
noise processes, $\boldsymbol{\nu}_{ba}$ and $\boldsymbol{\nu}_{bg}$, of power spectral
densities, $\sigma^2_{ba}\mathbf{I}_3$ and $\sigma^2_{bg}\mathbf{I}_3$,
respectively.
For brevity, we denote the IMU readings from $t_i$ to $t_j$ by $\mathbf{u}_{i:j} = \{(\boldsymbol{\omega}_m, \mathbf{a}_m)_k | k = i, i+1, \dots, j\}$

With a sequence of IMU readings $\mathbf{u}_{i:j}$,
the navigation state variable $\mathbf{x}(t_j)$ can be propagated from $\mathbf{x}(t_i)$ as expressed by $\mathbf{f}(\cdot)$,
\begin{equation}\label{eq:imu_prop}
\mathbf{x}(t_j | t_i) =
\mathbf{f}(\mathbf{x}(t_i), \mathbf{u}_{i:j},
\mathbf w_{imu}),
\end{equation}
where the continuous noises of IMU readings are stacked in $\mathbf w_{imu} = [\boldsymbol{\nu}_g^\intercal, \boldsymbol{\nu}_a^\intercal, 
\boldsymbol{\nu}_{bg}^\intercal, \boldsymbol{\nu}_{ba}^\intercal]^\intercal$.
For brevity, we will drop the time symbol and keep only its index for variables in \eqref{eq:imu_prop},
\eg, $\mathbf{x}_{j|i} = \mathbf{x}(t_j | t_i)$. 
The propagated navigation state $\boldsymbol{\pi}_{j|i}$ can be solved with the Runge-Kutta method \cite{jekeli_inertial_2001}.

\subsection{Global Bundle Adjustment and FLS}
Before looking at the FLS, we first presents the basics of
global bundle adjustment (BA) \cite{triggsBundle2000} which is the base of the FLS.
For the visual inertial SLAM problem, the objective function to be minimized in the global BA up to $t_k$ is
\begin{equation}
\label{eq:energy_function}
\begin{split}
E &= \sum_{i=1}^k\Vert \mathbf{r}_x(\mathbf{x}_{i},  \mathbf{x}_{i|i-1})\Vert^2_{\mathbf{\Sigma}_{x, i-1:i}} + \\
&\quad \sum_{(i, l) \in \mathcal{C}_k} \Vert \mathbf{r}_{il}(\mathbf{x}_i, \mathbf{x}_a, \mathbf{f}_l) \Vert^2_{\boldsymbol{\Sigma}_{c}},
\end{split}
\end{equation}
where $\mathbf{r}_x$ and $\mathbf{r}_{il}$ are residual errors associated with IMU and camera measurements, 
$\mathbf{\Sigma}_{x, k-1:k}$ and $\mathbf{\Sigma}_c$ are their corresponding covariance matrices,
and $\mathcal{C}_k$ denotes all image measurements up to $t_k$.
Note that the objective function does not include a gauge-fixing prior
which will shadow unobservable directions.

The reprojection error $\mathbf{r}_{il}$ is usually defined to be the
mismatch between predicted image coordinates of a landmark $\mathbf{f}_l$ and its measurement $\mathbf{z}_{il}$, \ie,
$\mathbf{r}_{il} = \mathbf{h}(\mathbf{x}_i,	\mathbf{x}_a, \mathbf{f}_l) - \mathbf{z}_{il}$.

The IMU residual error $\mathbf{r}_x$ and its covariance depends on the error definitions and will be discussed later on.

Solving the objective function \eqref{eq:energy_function} is equivalent to finding a solution to fit the below nonlinear system,
\begin{equation}
\label{eq:nonlinear-fit}
\begin{split}
\mathbf{W}\underbrace{\left[\begin{array}{c}
	\mathbf{r}_x(\mathbf{x}_{1},  \mathbf{x}_{1|0})\\ 
	\vdots \\ 
	\mathbf{r}_x(\mathbf{x}_{k},  \mathbf{x}_{k|k-1})\\ 
	\hline
	\vdots \\ 
	\mathbf{r}_{il}(\mathbf{x}_i, \mathbf{x}_a, \mathbf{f}_l) \\
	\vdots
	\end{array}\right]}_\mathbf{r} = \mathbf{0} \\
\mathbf{W} = \left[\begin{array}{c|c}
	\begin{matrix} \mathbf{\Sigma}_{x, 0:1}^{-1/2} &  & \\
	&  \ddots &  \\
	&  &  \mathbf{\Sigma}_{x, k-1:k}^{-1/2}
	\end{matrix} & \begin{matrix}
	&  &  & \\
	&  &  & \\ 
	&  &  & 
	\end{matrix} \\ 
	\hline
	\begin{matrix}
	&  &  & \\
	&  &  & \\
	&  &  & 
	\end{matrix} &
	\begin{matrix}
	\mathbf{\Sigma}_{c, 1}^{-1/2} &  & \\ 
	 & \ddots & \\ 
	 &  & \mathbf{\Sigma}_{c, m}^{-1/2}
	\end{matrix}
	\end{array}\right]
	\end{split}
\end{equation}
where $m = |\mathcal{C}_k|$ is the total number of image observations.

FLS minimizes \eqref{eq:energy_function} by repeatedly going through two steps, linearization and marginalization, as described next.
\subsubsection{Factor Linearization}
Before linearizing the measurement factors, the error state (\ie, the `small' perturbation) must be defined.
Without loss of generality, we define the error state $\delta \mathbf{x}$ as a function
$\boldsymbol{\eta}$ of the random variable $\mathbf{x}$ and its noise free estimate $\bar{\mathbf{x}}$, \ie,
$\delta \mathbf{x} = \boldsymbol{\eta}(\mathbf{x}, \bar{\mathbf{x}})$.
For a variable in a real vector space, the error state is simply $\delta \mathbf{x} = \mathbf{x} - \bar{\mathbf{x}}$.
Also, we define the inverse of $\boldsymbol{\eta}$ such that
$\mathbf{x} = \boldsymbol{\eta}^{-1}(\bar{\mathbf{x}}, \delta \mathbf{x})$.

With the error state, the residual errors can be linearized at estimates of state variables with the first order approximation.
The reprojection error is linearized as
\begin{equation}
\begin{split}
\mathbf{r}_{il}(\mathbf{x}_i, \mathbf{x}_a, \mathbf{f}_l) &\approx \mathbf{r}_{il}(\bar{\mathbf{x}}_i, \bar{\mathbf{x}}_a, \bar{\mathbf{f}}_l) +
\mathbf{J}_{x_i, l} \delta \mathbf{x}_i + \\
& \quad \mathbf{J}_{x_a, l} \delta \mathbf{x}_a + \mathbf{J}_{f_l} \delta \mathbf{f}_l
\end{split}
\end{equation}
where $\mathbf{J}_{x_i, l}$, $\mathbf{J}_{x_a, l}$, and $\mathbf{J}_{f_l}$ are Jacobians of $\mathbf{r}_{il}$
relative to $\mathbf{x}_i$, $\mathbf{x}_a$, and $\mathbf{f}_l$.

The above-mentioned IMU residual error $\mathbf{r}_x$ is usually defined to be
$\mathbf{r}_x(\mathbf{x}_i, \mathbf{x}_{i|i-1}) = \boldsymbol{\eta}(\mathbf{x}_i, \mathbf{x}_{i|i-1})$.
It is linearized as
\begin{equation}
\label{eq:imu-residual-lin}
\begin{split}
\mathbf{r}_x(\mathbf{x}_{i}, \mathbf{x}_{i|i-1}) &\approx 
\mathbf{r}_x(\bar{\mathbf{x}}_{i},  \bar{\mathbf{x}}_{i|i-1}) + \mathbf{A}_{i} \delta \mathbf{x}_i + \\
& \quad \mathbf{A}_{i|i-1} \delta \mathbf{x}_{i|i-1} \\
&= \mathbf{r}_x(\bar{\mathbf{x}}_{i},  \bar{\mathbf{x}}_{i|i-1}) + \mathbf{A}_{i} \delta \mathbf{x}_i + \\
& \quad \mathbf{A}_{i|i-1} \boldsymbol{\Phi}(t_i, t_{i-1}) \delta \mathbf{x}_{i-1}
\end{split}
\end{equation}
where $\mathbf{A}_{i}$ and $\mathbf{A}_{i|i-1}$ are the Jacobians of $\mathbf{r}_x$
relative to $\delta \mathbf{x}_i$ and $\delta \mathbf{x}_{i|i-1}$, and $\boldsymbol{\Phi}(t_i, t_{i-1})$ is
the discrete IMU transition matrix obtained by linearizing \eqref{eq:imu_prop}.
To obtain the weight covariance $\mathbf{\Sigma}_{x, i-1:i}$, we note that the covariance of $\delta \mathbf{x}_{i|i-1}$,
$\mathbf{\Sigma}_{\mathbf{x}_{i|i-1}}$, can be propagated from a zero matrix 
by the propagation function \eqref{eq:imu_prop} given the defined error state $\delta \mathbf{x}$,
then $\mathbf{\Sigma}_{x, i-1:i} =  \mathbf{A}_{i|i-1} \mathbf{\Sigma}_{\mathbf{x}_{i|i-1}} \mathbf{A}_{i|i-1}^\intercal$.

Linearization turns the nonlinear system \eqref{eq:nonlinear-fit} to a set of linear equations that we try to satisfy at once,
\begin{equation}
\label{eq:linear-fit}
\begin{split}
\mathbf{W}\left(\underbrace{\left[\begin{array}{c}
	\mathbf{r}_x(\bar{\mathbf{x}}_{1},  \bar{\mathbf{x}}_{1|0})\\ 
	\vdots \\ 
	\mathbf{r}_x(\bar{\mathbf{x}}_{k},  \bar{\mathbf{x}}_{k|k-1})\\ 
	\hline
	\vdots \\ 
	\mathbf{r}_{il}(\bar{\mathbf{x}}_i, \bar{\mathbf{x}}_a, \bar{\mathbf{f}}_l) \\
	\vdots
	\end{array}\right]}_{\bar{\mathbf{r}}} +
\mathbf{J} 
\underbrace{
\left[\begin{array}{c}
\delta \mathbf{x}_0 \\
\vdots \\
\delta \mathbf{x}_k \\
\hline
\vdots \\
\delta \mathbf{f}_l \\
\vdots
\end{array}\right]}_{\delta \mathcal{X}_k}\right) =\mathbf{0} \\
\mathbf{J} = 
\left[\begin{array}{c|c}
\begin{matrix}
\mathbf{A}_{1|0} \boldsymbol{\Phi}_{1|0} &  \mathbf{A}_1 & &\\
&\dots&&\\
&& \mathbf{A}_{k|k-1} \boldsymbol{\Phi}_{k|k-1} &  \mathbf{A}_k
\end{matrix} &
\begin{matrix}
& & \\
& & \\
& &
\end{matrix} \\ \hline
\begin{matrix}
&&\dots&& \\
&\mathbf{J}_{x_i, l}& & \mathbf{J}_{x_a, l}& \\
&&\dots&&
\end{matrix} & 
\begin{matrix}
&\dots& \\
&\mathbf{J}_{f_l}& \\
&\dots&
\end{matrix}
\end{array}
\right].
\end{split}
\end{equation}

\subsubsection{Unobservable Directions and Nullspace}
To analyze the observability of the visual inertial SLAM problem, we need to specify the unobservable directions and 
relate them to the objective function \eqref{eq:energy_function} and the linearized system \eqref{eq:linear-fit}.

First, let's define a transformation $\mathcal{T}_\xi$ (minimally parameterized by $\boldsymbol{\xi}$) of the considered problem which
transforms all state variables from the present world frame \{$W$\} to another one, say \{$W_y$\}.
The transformation applies to all variables relevant to \{$W$\}, \ie, $\mathbf{x}_{0:k}$,
which become $\mathbf{y}_{0:k}$ after the transformation.

With measurements from a camera and a consumer-grade IMU, it is
impossible to determine the absolute position and heading of the platform \cite{jones2011visual} which are the unobservable directions for the visual inertial SLAM problem.
When a transformation involves only an translation 
$\delta \mathbf{t}$ and a rotation about gravity
$\delta \phi$, \ie, $\boldsymbol{\xi} = [\delta \phi \enskip \delta \mathbf{t}]$,
the value of the objective function \eqref{eq:energy_function} is invariant to the transformation because the residual errors do not change w.r.t the new variables $\mathbf{y}_{0:k}$, \ie,
\begin{equation}
\begin{split}
\mathbf{r}_x(\mathbf{x}_{i},  \mathbf{x}_{i|i-1}) = \mathbf{r}_x(\mathcal{T}_\xi(\mathbf{x}_{i}),  \mathcal{T}_\xi(\mathbf{x}_{i|i-1}))\\
\mathbf{r}_{il}(\mathbf{x}_i, \mathbf{x}_a, \mathbf{f}_l) =
\mathbf{r}_{il}(\mathcal{T}_\xi(\mathbf{x}_i), \mathcal{T}_\xi(\mathbf{x}_a), \mathbf{f}_l)
\end{split}	
\end{equation}
Thus, the linearized system \eqref{eq:linear-fit} still holds but with $\mathbf{y}_{0:k}$.

Next, we reveal that the unobservable directions correspond to the nullspace of $\mathbf{J}$ in \eqref{eq:linear-fit}. 
When $\boldsymbol{\xi}$ is close to the zero vector, the objective function after the transformation $\mathcal{T}_\xi$ 
can be linearized at the estimates for $\mathcal{X}_{k}$, and the linearized system becomes
\begin{equation}
\label{eq:linear-fit-T}
\mathbf{W}(\bar{\mathbf{r}} + \mathbf{J} \delta \mathcal{Y}_k) = \mathbf{0}
\end{equation}
where $\delta \mathcal{Y}_k$ is the error between the transformed state variables $\mathcal{Y}_k = \mathcal{T}_\xi(\mathcal{X}_k)$ and their linearization points $\bar{\mathcal{X}}_k$.
By comparing \eqref{eq:linear-fit} and \eqref{eq:linear-fit-T}, we observe that 
\begin{equation}
\begin{split}
\mathbf{0} &= \mathbf{J} (\delta \mathcal{Y}_k - \delta \mathcal{X}_k) \\ 
&= \mathbf{J} (\boldsymbol{\eta}(\mathcal{T}_\xi(\mathcal{X}_k), \bar{\mathcal{X}}_k) - \boldsymbol{\eta}(\mathcal{X}_k, \bar{\mathcal{X}}_k)) \\
&= \mathbf{J} \underbrace{\frac{\partial \boldsymbol{\eta}(\mathcal{T}_\xi(\mathcal{X}_k), \bar{\mathcal{X}}_k)}{\partial \boldsymbol{\xi}}}_{\mathbf{N}_J} \boldsymbol{\xi}.
\end{split}
\end{equation}
As the expression holds for arbitrary small $\boldsymbol{\xi}$, we have 
\begin{equation}
	\mathbf{J}\mathbf N_{J} = \mathbf{0},
\end{equation}
which means that changes on the column space of $\mathbf{N}_J$ to variables do not affect the linearized system.
Thus, the nullspace of $\mathbf{J}$, $\mathbf{N}_J$ corresponds to the unobservable directions of the problem.
In this sense, the Jacobian matrix of the system $\mathbf{J}$ is
equivalent to the classic observability matrix in revealing the unobservable directions.
Indeed, the observability matrix can be obtained from $\mathbf{J}$ by
basic row operations as shown in \cite{dong-siMotionTrackingFixedlag2011}.

\subsubsection{Factor Marginalization}
The FLS has been a popular approach to the visual inertial SLAM problem, \eg, \cite{rosinolKimera2020}. 
Essentially, it solves the problem by repeated linearization of  factors,
and gradually marginalizes old variables from the global BA problem to bound problem size.
Every marginalization step creates a linear prior factor for variables connected to those removed variables.

Consider a marginalization step where variables prior to $t_m$ are marginalized.
The objective function \eqref{eq:energy_function} becomes
\begin{equation}
\begin{split}
	E_m &= \sum_{i=1}^m\Vert 
	\mathbf{r}_x(\bar{\mathbf{x}}_{i}, \bar{\mathbf{x}}_{i|i-1}) + \mathbf{A}_{i} \delta \mathbf{x}_i + \\
	&\quad \mathbf{A}_{i|i-1} \boldsymbol{\Phi}(t_i, t_{i-1}) \delta \mathbf{x}_{i-1}
	 \Vert^2_{\mathbf{\Sigma}_{x, i-1:i}} + \\
	&\quad \sum_{(i, l) \in \mathcal{M}} \Vert 
	\mathbf{r}_{il}(\bar{\mathbf{x}}_i, \bar{\mathbf{x}}_a, \bar{\mathbf{f}}_l) + \mathbf{J}_{x_i, l} \delta \mathbf{x}_i +\\
	&\quad \mathbf{J}_{x_a, l} \delta \mathbf{x}_a + \mathbf{J}_{f_l} \delta \mathbf{f}_l
	 \Vert^2_{\boldsymbol{\Sigma}_{c}} + \\
	&\quad \sum_{i=m+1}^k\Vert \mathbf{r}_x(\mathbf{x}_{i}, \mathbf{x}_{i|i-1})\Vert^2_{\mathbf{\Sigma}_{x, i-1:i}} + \\
	&\quad \sum_{(i, l) \in \mathcal{C}_k \setminus \mathcal{M}} \Vert \mathbf{r}_{il}(\mathbf{x}_i, \mathbf{x}_a, \mathbf{f}_l) \Vert^2_{\boldsymbol{\Sigma}_{c}},
\end{split}
\end{equation}
where $\mathcal{M}$ is the set of marginalized camera observations.
The first two linear terms of $E_m$ are
obtained by fixing linearization points for the marginalized nonlinear factors.
For convenience of analysis, none of the marginalized terms is discarded.
In implementation, the first two linear terms of $E_m$ are equivalently
represented by a marginalization factor which is obtained by 
the Schur complement method.

As the optimizer iterates, $E_m$ will be relinearized. 
For a variable in the marginalization factor (\eg, $\mathbf{x}_{m}$),
a nonlinear term of $E_m$ usually will be linearized at a different estimate ($\bar{\mathbf{x}}^\prime_{m}$)
than the one ($\bar{\mathbf{x}}_{m}$) used in the linear terms of $E_m$.
Thus, the Jacobian matrix in the linearized system \eqref{eq:linear-fit} will 
have blocks evaluated at different points for the same variables in the marginalization factor.
For the traditional error definition,
this causes shrunk nullspace of $\mathbf{J}$ and inconsistent covariances as shown in \cite{dong-siMotionTrackingFixedlag2011}.

\subsection{Right Invariant Fixed-Lag Smoother}
\label{subsec:ri-fls}
In contrast to traditional error definitions, the right invariant error formulation 
does not suffer from this inconsistency in observable dimensions and covariances.

\subsubsection{The Right Invariant Error}
The right invariant error is defined relative to the navigation state $\boldsymbol{\pi}_i$, viewed as an element $X_i$ of $SE_2(3)$ \cite{barrauInvariantExtendedKalman2016}, \ie,
\begin{equation}
X_i = \begin{bmatrix}
\mathbf{R}_i & \mathbf{v}_i & \mathbf{p}_i\\
\mathbf{0}_{1\times3} & 1 & 0\\
\mathbf{0}_{1\times3} & 0 & 1
\end{bmatrix} \in SE_2(3).
\end{equation}
The right invariant error $\boldsymbol{\xi}_{\pi, i}$ consisting of rotational error $\delta \boldsymbol{\theta}_i$, 
velocity error $\delta \mathbf{v}_i$, and positional error $\delta \mathbf{p}_i$, is given by
\begin{align}
\boldsymbol{\xi}_{\pi, i} &= (\delta \boldsymbol{\theta}_i, \delta \mathbf{v}_i, \delta \mathbf{p}_i), \\
X_i &= \exp(\mathcal{L}(\boldsymbol{\xi}_{\pi, i}))\bar{X}_i,
\label{eq:ri_relation}, \\
\mathcal{L}(\boldsymbol{\xi}_\pi) &= \begin{bmatrix}
\delta \boldsymbol{\theta}_\times & \delta \mathbf{v} & \delta \mathbf{p} \\
\mathbf{0}_{1\times 3} & 1 & 0 \\
\mathbf{0}_{1\times 3} & 0 & 1
\end{bmatrix}
\end{align}
where $\mathrm{exp}(\cdot)$ is the matrix exponent, and $\mathcal{L}(\boldsymbol{\xi}_\pi)$ is the Lie operator for $SE_2(3)$, computed with the skew operator $(\cdot)_\times$.
The closed form expression for the exponential map of $\boldsymbol{\xi}_\pi$ is,
\begin{equation}
\begin{split}
\exp(\mathcal{L}(\boldsymbol{\xi}_\pi)) = \begin{bmatrix}
\exp(\delta \boldsymbol{\theta}_\times) & \mathbf{J}_l(\delta \boldsymbol{\theta}) \delta \mathbf{v} & 
\mathbf{J}_l(\delta \boldsymbol{\theta}) \delta \mathbf{p} \\
\mathbf{0}_{1\times3} & 1 & 0 \\
\mathbf{0}_{1\times3} & 0 & 1
\end{bmatrix}
\end{split}
\end{equation}
where $\mathbf{J}_l(\cdot)$ is the left Jacobian for $SO(3)$ \cite{barfootAssociating2014}.

``Right invariance" is on the grounds that
the error for $X$ is the same as that for its transformed variable, $XY$, 
obtained by right multiplication with an element $Y\in SE_2(3)$, as shown by
$X Y = \exp(\mathcal{L}(\boldsymbol{\xi}_{\pi}))(\bar{X} Y)$.
That is, the right invariant error is independent of the system state.

\subsubsection{Consistency Property}
For the right invariant errors, assuming $\Delta t = t_i - t_{i-1}$ is small (\eg, 0.1s), 
the discrete transition matrix $\boldsymbol{\Phi}(t_i, t_{i-1})$ is found to be
\begin{equation}
\label{eq:coarse-phi}
\begin{split}
&\boldsymbol{\Phi}_{i|i-1}
=\begin{bmatrix}
\boldsymbol{\Phi}_\pi & \boldsymbol{\Phi}_{\pi, b} \\
\mathbf{0} & \mathbf{I}
\end{bmatrix} \\
\boldsymbol{\Phi}_\pi &= 
\begin{bmatrix}
\mathbf{I} & \mathbf{0} & \mathbf{0}\\
\mathbf{g}_\times\Delta t & \mathbf{I} & \mathbf{0}\\
\mathbf{g}_\times\Delta t^2/2 & \mathbf{I}\Delta t & \mathbf{I}
\end{bmatrix} \\
\boldsymbol{\Phi}_{\pi, b}
& = \begin{bmatrix}
-\mathbf{R}\Delta t & \mathbf{0} \\ 
- \mathbf{v}_\times \mathbf{R}\Delta t - \mathbf{g}_\times \mathbf{R}\frac{\Delta t^2}{2} & -\mathbf{R}\Delta t \\
- \mathbf{p}_\times \mathbf{R}\Delta t - \mathbf{v}_\times \mathbf{R} \frac{\Delta t^2}{2} - \mathbf{g}_\times \mathbf{R}\frac{\Delta t^3}{6} & -\mathbf{R}\frac{\Delta t^2}{2}
\end{bmatrix}
\end{split}
\end{equation}
where $\mathbf{g}$ is the gravity vector in \{$W$\}, 
and we drop the subscript `i' of $(\mathbf{R, v, p})$ for brevity.
Thanks to right invariance, $\boldsymbol{\Phi}_\pi$ is independent of the state variable $\boldsymbol{\pi}_i$. 

Another useful finding is that the parameters of landmarks anchored at a camera frame and 
sensor parameters (\eg, biases) do not interfere with nullspace of
the coefficient matrix $\mathbf{J}$ of the linearized system \eqref{eq:linear-fit}.
Thus, their Jacobians can be safely ignored in analyzing consistency.

The right invariance property together with the above finding lead to the proof that 
the right invariant error formulation can ensure consistency of the
FLS as detailed in the supplementary material.

One point worth noting is that the proof approximates two component Jacobians for the IMU residual error \eqref{eq:imu-residual-lin},
$\mathbf{A}_{i}$ and $\mathbf{A}_{i|i-1}$, by identities,
\begin{equation}
\label{eq:imu_jac_approx}
	\mathbf{A}_{i} \approx \mathbf{I}_{15} \quad \mathbf{A}_{i|i-1} \approx -\mathbf{I}_{15}.
\end{equation}
This approximation is also used in \cite{dong-siMotionTrackingFixedlag2011} for proving consistency of the FEJ technique.
It is reasonable when the IMU residual error $\mathbf{r}_{x}$ is small,
and we found that using the exact $\mathbf{A}_{i}$ and $\mathbf{A}_{i|i-1}$ led to slight inconsistency (see Fig.~\ref{fig:nees-rifls}).

\section{Simulation Results}
\label{sec:simulation}
This section presents the simulation results, validating that the FLS formulated with right invariant errors has consistent covariances.

\subsubsection{Error Metrics} The consistency of a FLS is measured by the Normalized Estimation
Error Squared (NEES) of components of the navigation state variable.
The expected value of NEES for a variable is its degrees of freedom, 
hence 3 for positional error $\delta \mathbf{p}_{WB}$, 
3 for orientation error $\delta \boldsymbol{\theta}_{WB}$, 
and 6 for pose error $\delta\mathbf{T}_{WB} = (\delta \mathbf{p}_{WB},\delta \boldsymbol{\theta}_{WB})$.
An inconsistent estimator will optimistically estimate the covariance,
thus the computed NEES is greater than its expected value.
Following \cite[(3.7.6-1)]{bar-shalomEstimationApplicationsTracking2004}, with $n_s$ successful runs of an estimator,
the NEES $\boldsymbol{\epsilon}$ for position, orientation, and pose at epoch $t$ is given by
\begin{align}
\boldsymbol{\epsilon}_{X}(t) = \frac{1}{n_s}\Sigma_{i=1}^{n_s} \delta
\mathbf{X}(t)^\intercal \boldsymbol{\Sigma}_{X}^{-1}(t) \delta \mathbf{X}(t)
\end{align}
where $\mathbf{X} = \mathbf{p}_{WB}, \boldsymbol{\theta}_{WB}, \mathbf{T}_{WB}$,
and $\boldsymbol{\Sigma}_{X}$ is its covariance.

The accuracy of the estimated state is measured by Root Mean Square Error (RMSE) for components of the state vector.
A component $\mathbf{X}$'s RMSE $r_X$ at $t$ is given by 
\begin{equation}
r_X(t) = \sqrt{\frac{1}{n_s}\Sigma_{i=1}^{n_s} \delta \mathbf{X}(t)^{\intercal} \delta \mathbf{X}(t)}
\end{equation}

\subsubsection{Simulation Setup}
\begin{figure}[]
	\centering
	\includegraphics[width=0.85\columnwidth]{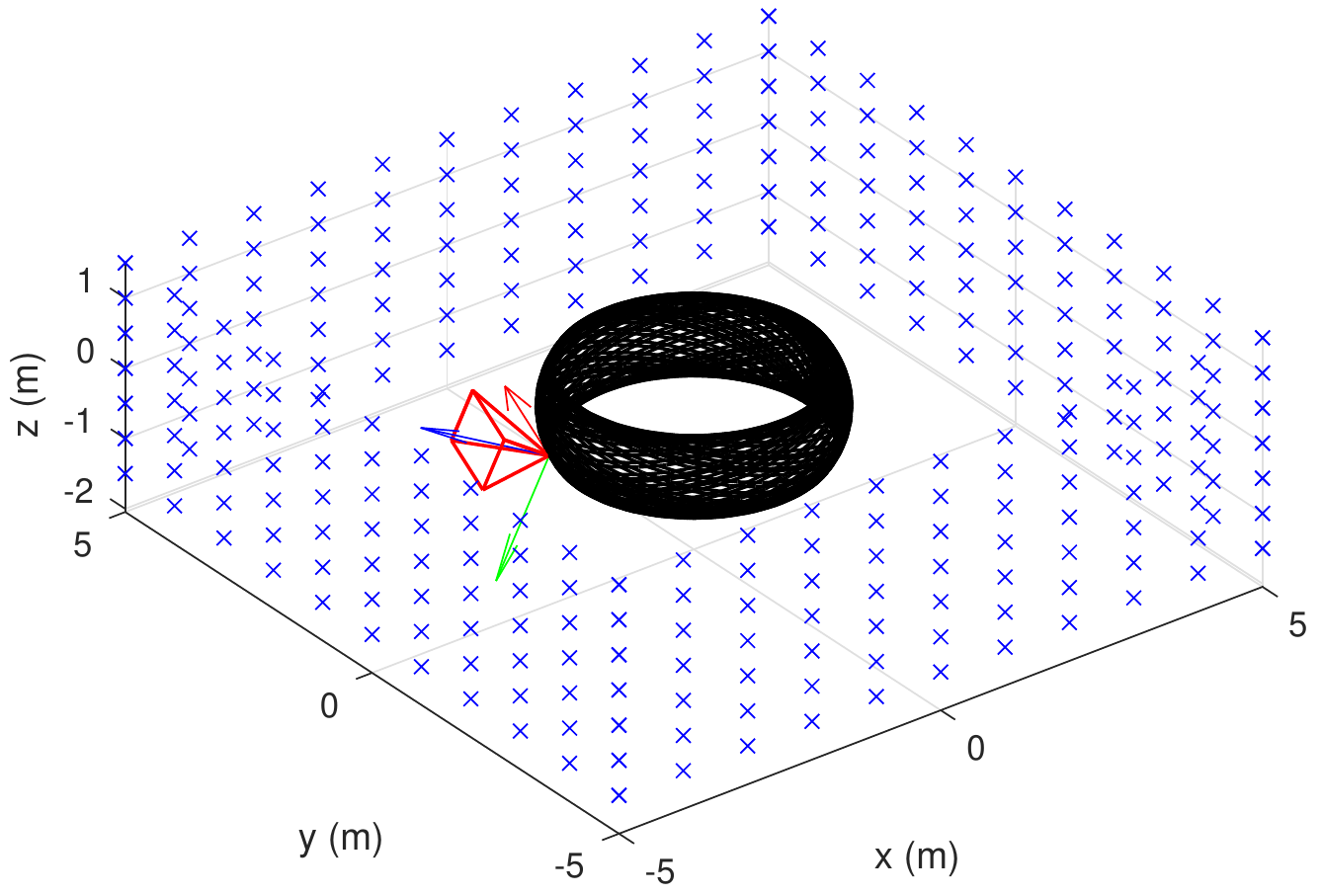}
	\caption{Simulated scene with general torus motion lasting for five minutes. A sample camera view frustum is shown by the red wireframe on the black trajectory.}
	\label{fig:scenarios}
\end{figure}

A scene with point landmarks distributed on four walls was simulated.
A monocular camera-IMU platform traversed the scene for five minutes
with a torus trajectory (Fig.~\ref{fig:scenarios}).
The platform moved at an average velocity 2.30 m/s.

\begin{table}[]
	\centering
	\begin{tabular}{lll}
		\hline
		$\Sigma$ & Gyroscope  & Accelerometer \\ \hline
		\begin{tabular}[c]{@{}l@{}}Bias White\\ Noise \\ 
		\end{tabular} &
		\begin{tabular}[c]{@{}l@{}}$\sigma_{bg}^2 / f \mathbf{I}_3$ with\\
			$\sigma_{bg}=2\cdot10^{-5}$\\ $rad/s^2/\sqrt{Hz}$\end{tabular} &
		\begin{tabular}[c]{@{}l@{}}$\sigma_{ba}^2 / f \mathbf{I}_3$ with\\
			$\sigma_{ba} = 5.5\cdot10^{-5}$\\ $m/s^3/\sqrt{Hz}$\end{tabular} \\ \hline
		\begin{tabular}[c]{@{}l@{}}White Noise \\
		 \end{tabular}&
		\begin{tabular}[c]{@{}l@{}}$\sigma_g^2 f \mathbf{I}_3$ with\\
			$\sigma_g=1.2\cdot10^{-3}$\\ $rad/s/\sqrt{Hz}$\end{tabular}&
		\begin{tabular}[c]{@{}l@{}}$\sigma_a^2 f \mathbf{I}_3$ with\\ 
			$\sigma_a = 8\cdot10^{-3}$\\ $m/s^2/\sqrt{Hz}$\end{tabular} \\ \hline
	\end{tabular}
\caption{Covariances of the zero-mean Gaussian distributions from which
	discrete noise samples are drawn. $f$ is the IMU sampling rate.}
	\label{tab:imu_noise}
\end{table}

The camera captured images of size $752 \times 480$ at 10Hz.
The image observations were corrupted by white Gaussian noise of 1 pixel standard deviation at each direction.
The simulated inertial measurements were sampled at $f$=100 Hz,
corrupted by random walk biases and additive white noise.
Discrete noise samples were drawn from Gaussian distributions
tabulated in Table~\ref{tab:imu_noise}.
These noise parameters were chosen to be realistic for a consumer-grade IMU.

\subsubsection{Estimator Setup} 
The proposed FLS was implemented with the \texttt{IncrementalFixedLagSmoother} in
GTSAM \cite{dellaertFactorGraphsGTSAM2012} which wraps the iSAM2 \cite{kaess2012isam2} method.
By setting the time horizon to a large value, it turns into the iSAM2 
which gives results very close to a batch solution \cite{forster_manifold_2017}.
Also, GTSAM provides a \texttt{BatchFixedLagSmoother} wrapping a Levenberg-Marquardt solver 
which ensures consistency by locking variables in the marginalization factor.

We compared several estimators, the incremental FLS (Inc. FLS),
the batch FLS, iSAM2, and the proposed FLS with the right invariant error (RI-FLS).
The first three estimators used the error state defined in \cite{forster_manifold_2017}.
Except for iSAM2, the other estimators adopted a time horizon of 1 second.

A simulation frontend was created to provide feature tracks to an estimator.
It associated observations of a landmark between consecutive
frames and between current frame and a selected earlier reference frame.
For the torus motion, the average feature track length was 5.8,
and the average number of observed landmarks in an image was 40.5.

All estimators were initialized with the true pose but a noisy velocity estimate
affected by noise of Gaussian distribution 
$N(\mathbf{0}, 0.05^2 \mathbf{I}_3 \hspace{0.2em} m^2/s^4)$.
Each estimator ran 100 times, and only successful runs (with the error in position $\le$ 100 m at the end), were used 
to compute the error metrics.

\subsubsection{Estimator Consistency}
\begin{figure}[]
	\centering
	\includegraphics[width=0.8\columnwidth]{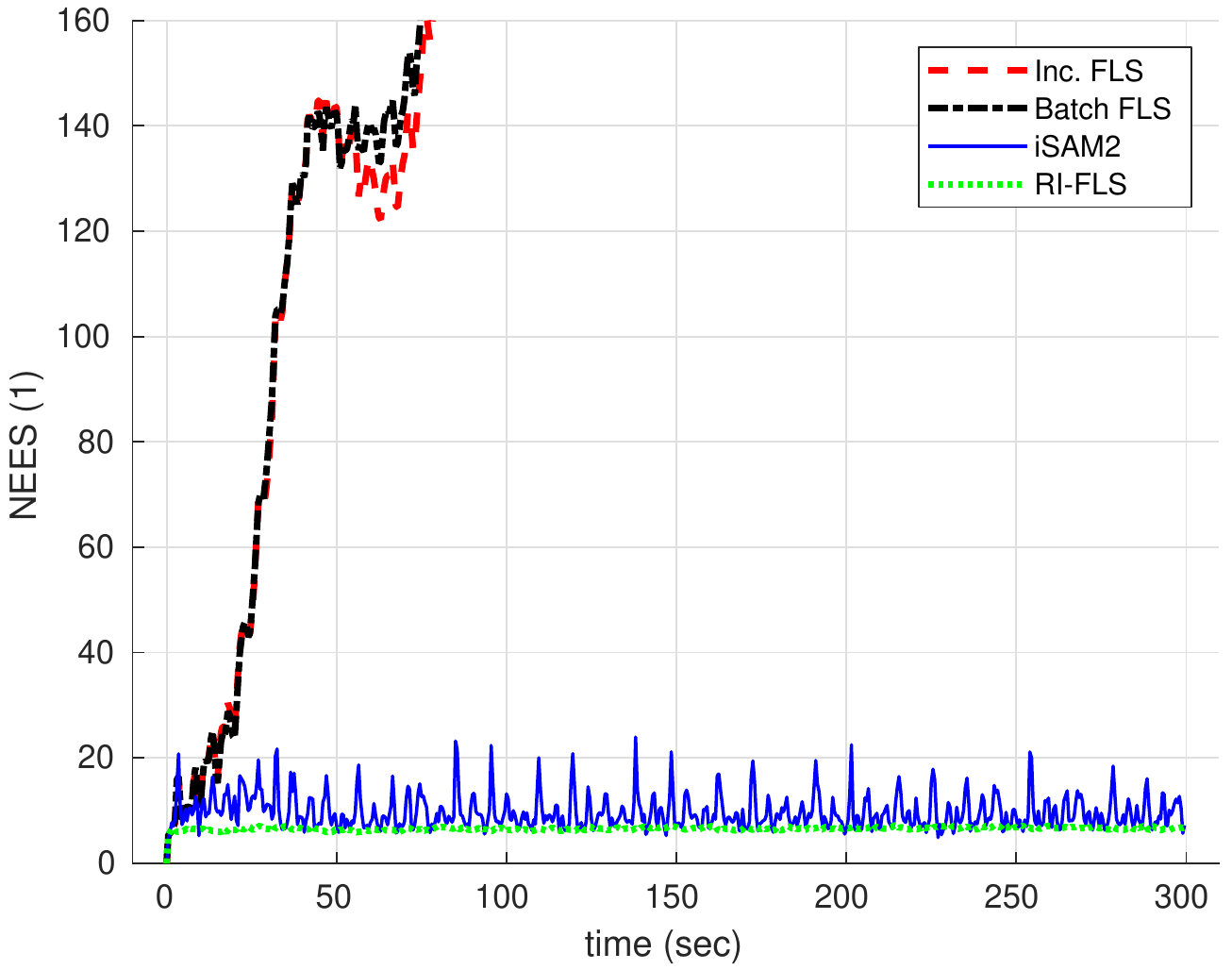}%
	\caption{The history of NEES for pose of
		estimators including incremental FLS, batch FLS, iSAM2, and right invariant FLS.
		The expected value of NEES for pose is 6.}
	\label{fig:nees}
\end{figure}

For the above estimators, the evolution of NEES is visualized in Fig.~\ref{fig:nees}. 
The NEES values averaged over the last 10 seconds to smooth out jitters are
tabulated in Table~\ref{tab:nees}.
From the NEES curves and their final values, we see that both incremental FLS and batch FLS did not
output consistent covariances, and incremental FLS performed better than batch FLS in terms of orientation NEES.
On the other hand, the proposed RI-FLS and iSAM2 achieved NEES values very close to the reference.
It is expected that iSAM2 exhibits consistency as it does not drop out variables.
It is a bit surprising that RI-FLS achieved even better NEES than iSAM2,
indicating that the right invariant formulation is effective for ensuring consistency.

\begin{figure}[]
	\centering
	\includegraphics[width=0.7\columnwidth]{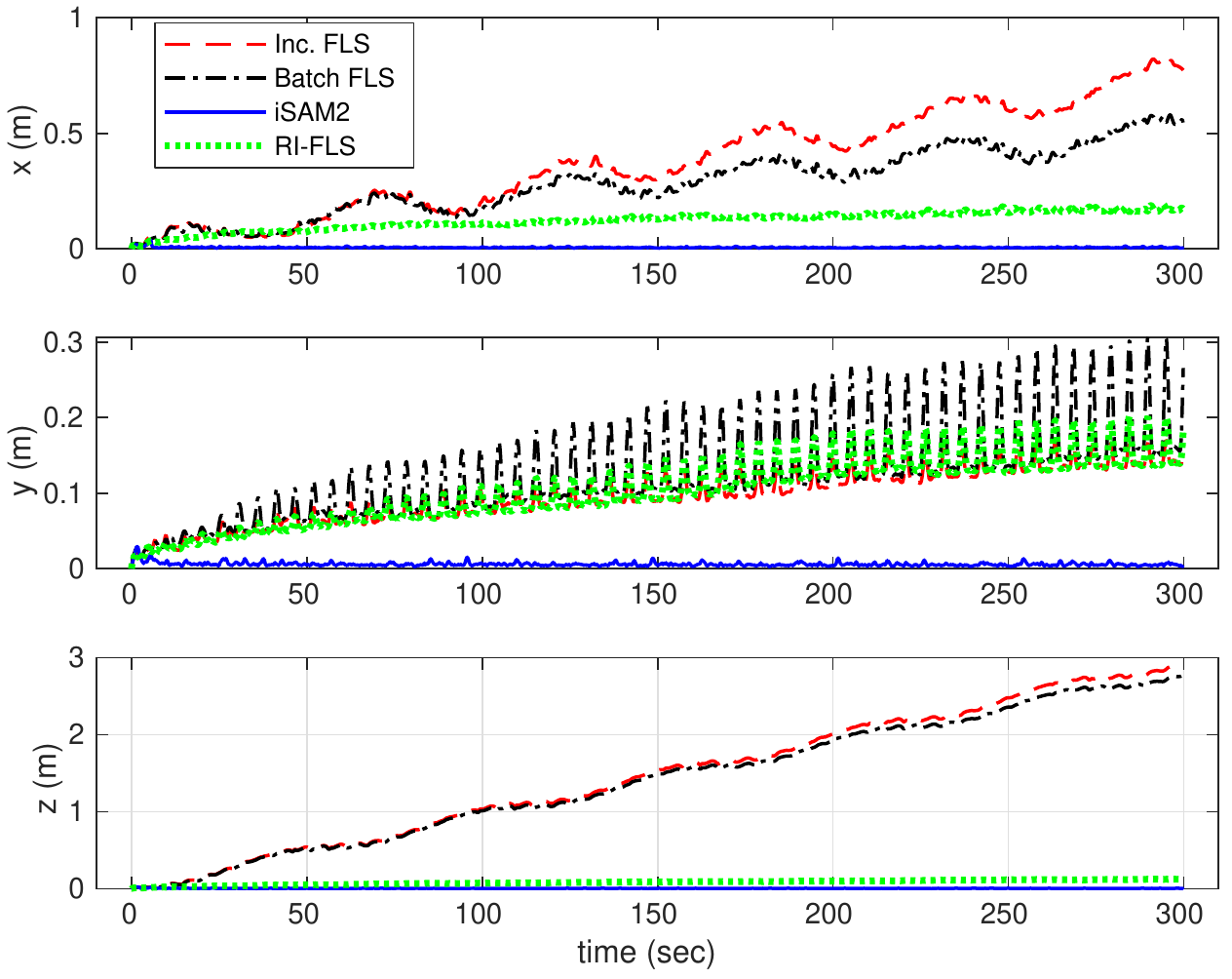} \\ (a) \\
	\includegraphics[width=0.7\columnwidth]{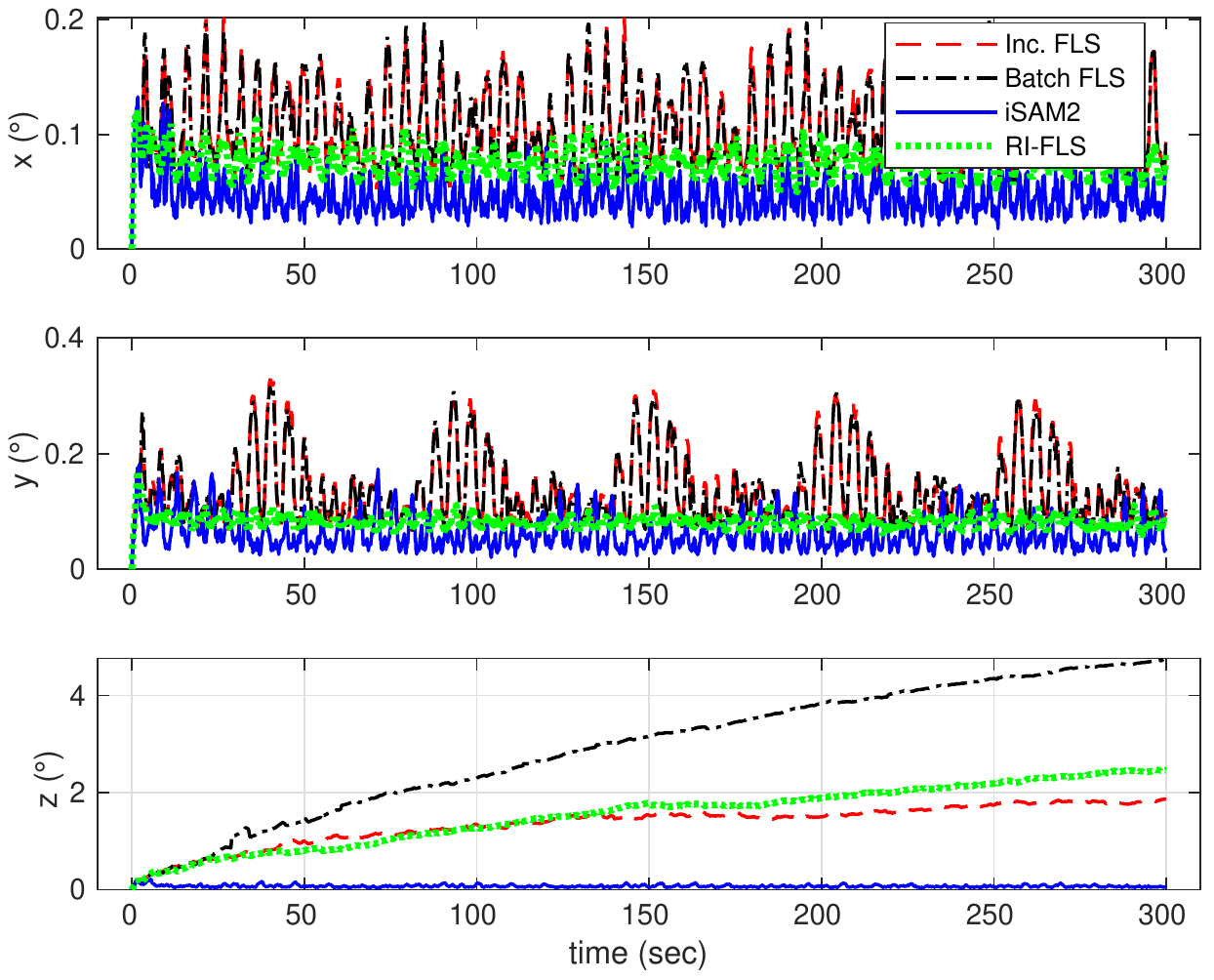} \\ (b) \\
	\includegraphics[width=0.7\columnwidth]{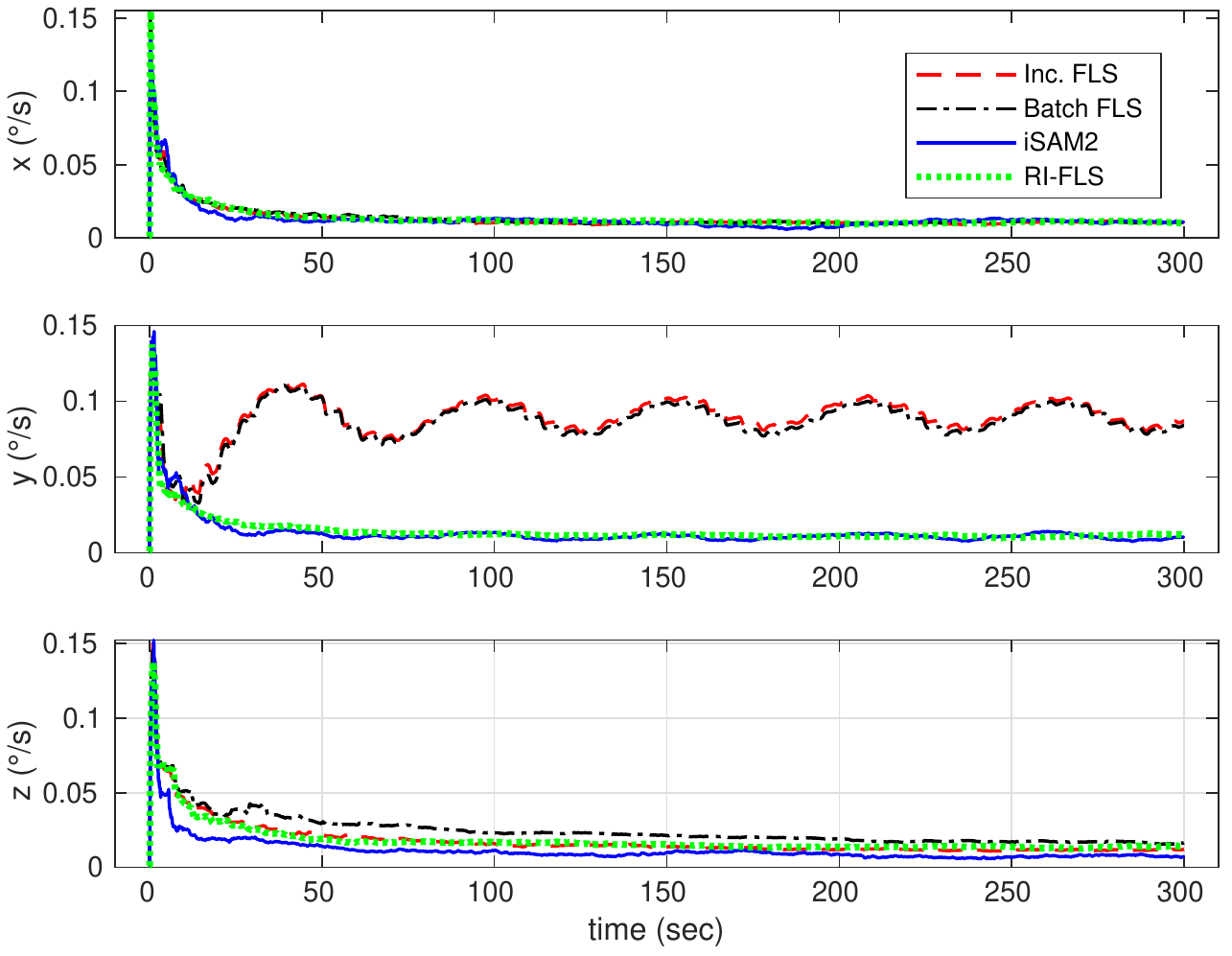} \\ (c) \\
	\includegraphics[width=0.7\columnwidth]{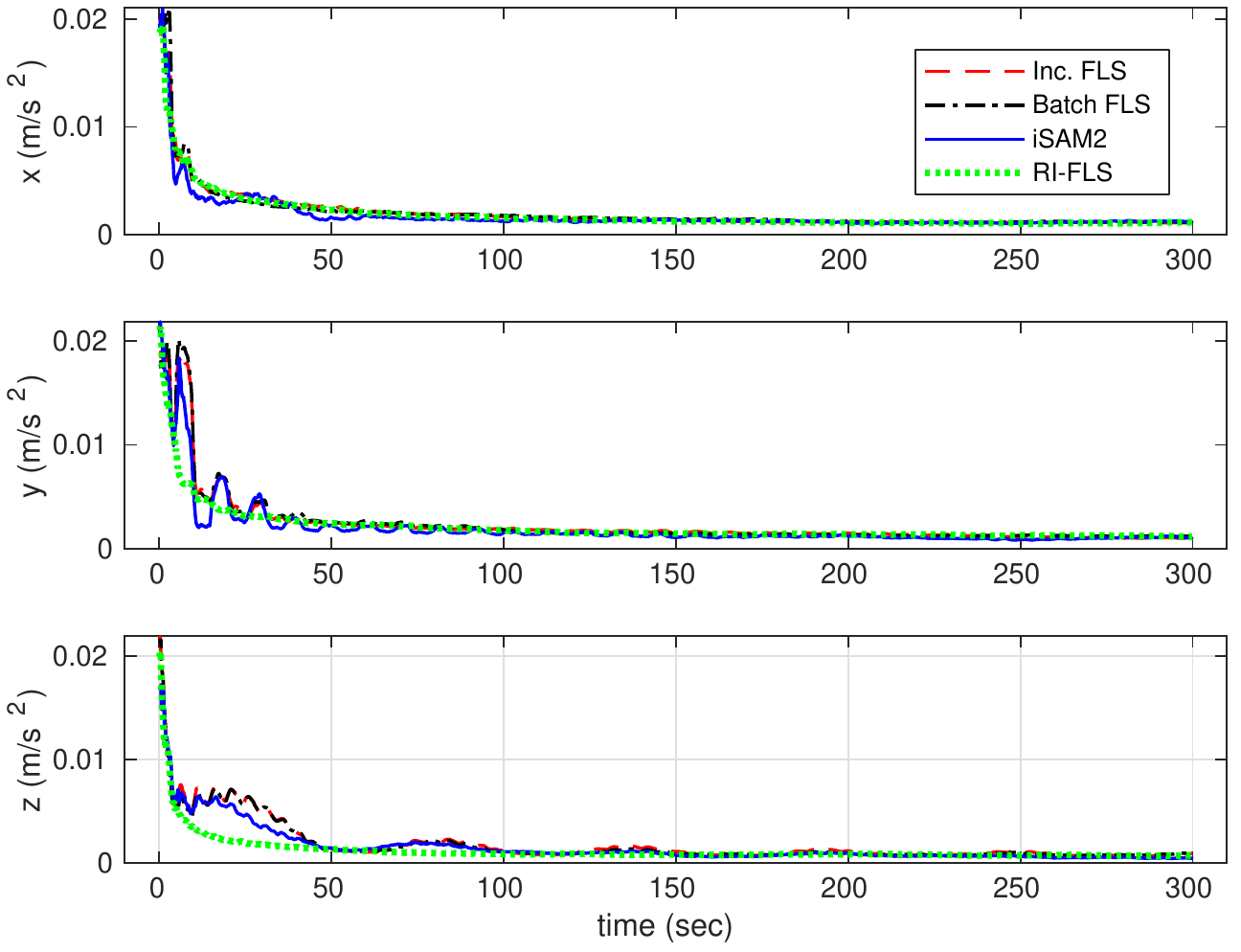} \\ (d)
	\caption{RMSE of position (a), orientation (b), gyro bias (c), and accelerometer bias (d), computed over 100
		runs for estimators including incremental FLS, batch FLS, iSAM2, and right invariant FLS.}
	\label{fig:rmse}
\end{figure}

To assess the state estimation accuracy, the RMSE values for each dimension of position, orientation, and IMU biases, are drawn in Fig.~\ref{fig:rmse}.
Unsurprisingly, iSAM2 achieved best accuracy for all these variables.
Incremental FLS and batch FLS had an issue in constraining errors on one horizontal direction of the gyro bias.
All estimators estimated well the accelerometer bias.
RI-FLS outperformed other FLSs in position accuracy, and achieved good orientation accuracy.

\begin{table}[]
	\begin{tabular}{l|lll|}
		\cline{2-4}
		& \multicolumn{3}{l|}{\begin{tabular}[c]{@{}l@{}}NEES averaged over last 10 seconds\end{tabular}} \\
		\cline{2-4}
		& \begin{tabular}[c]{@{}l@{}}Position (1)\end{tabular} & \begin{tabular}[c]{@{}l@{}}Orientation (1)\end{tabular}     
		& \begin{tabular}[c]{@{}l@{}}Pose (1)\end{tabular}\\ \hline
		\multicolumn{1}{|l|}{\begin{tabular}[c]{@{}l@{}}Reference\end{tabular}}
		& 3      & 3 & 6   \\ \hline
		\multicolumn{1}{|l|}{Inc. FLS}   & 644.3 &    5.0 & 657.3  \\ \hline
		\multicolumn{1}{|l|}{Batch FLS}     & 693.6  & 102.3    & 800.3  \\ \hline
		\multicolumn{1}{|l|}{iSAM2}       & 5.4 &    4.1 &  9.3  \\ \hline
		\multicolumn{1}{|l|}{RI-FLS}       &3.3 &    3.4 &    6.6 \\ \hline
	\end{tabular}
	\caption{NEES computed over 100 runs for estimators including incremental FLS, batch FLS, iSAM2, and right invariant FLS. }
	\label{tab:nees}
\end{table}

\subsubsection{RI-FLS variants} We also examine the effect of approximating the IMU residual Jacobians, 
and evaluate a RI-FLS variant with smart factors \cite{forster_manifold_2017}.

The consistency analysis approximates IMU residual Jacobians components $\mathbf{A}_i$ and $\mathbf{A}_{i|i-1}$ with identities \eqref{eq:imu_jac_approx}.
When the exact expressions for $\mathbf{A}_i$ and $\mathbf{A}_{i|i-1}$ are used, the observability property $\mathbf{JN}_J = \mathbf{0}$ may not hold.

The RI-FLS with smart factors is motivated by the fact that the GTSAM optimizer often throws the indeterminant system exception
because of landmarks with low disparity that are common for real data.
Smart factors fix this issue by removing landmarks from the optimizer.
We think this technique will not adversely impact estimator consistency.

\begin{figure}[]
	\centering
	\includegraphics[width=0.8\columnwidth]{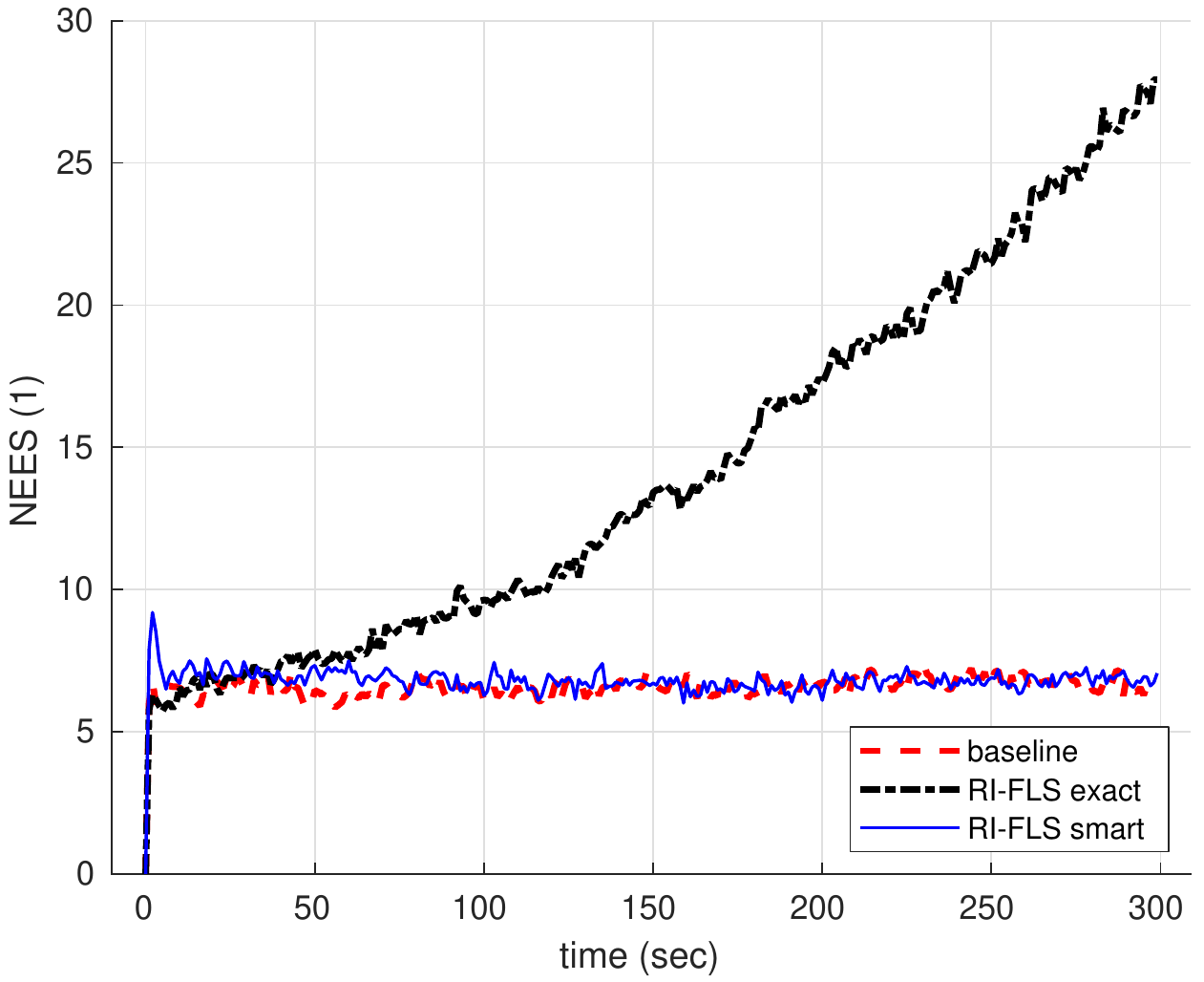}
	\caption{The history of NEES for pose of
		estimators including the baseline RI-FLS, RI-FLS with exact IMU Jacobians, and RI-FLS with smart factors.}
	\label{fig:nees-rifls}
\end{figure}

To confirm these thoughts, three variants of RI-FLS were tested in the above simulation setup: 
RI-FLS with approximated IMU Jacobians (baseline), RI-FLS with exact IMU Jacobians (RI-FLS exact),
and RI-FLS with smart factors and approximated IMU Jacobians (RI-FLS smart).
The history of the NEES for the three methods shown in Fig.~\ref{fig:nees-rifls} confirm that exact IMU Jacobians lead to worse NEES values,
and that smart factors do not worsen NEES values.

\section{Results on Real Data}
To show practicality, we tested the incremental FLS with errors defined in \cite{forster_manifold_2017}, 
RI-FLS, and RI-FLS with exact Jacobians on the EuRoC benchmark.
All methods were implemented with smart factors to handle degenerate landmarks and state variables were associated with consecutive camera frames in a time horizon of 1 second.
The absolute translation error RMS \cite{zhang2018tutorial} averaged over 3 runs on several EuRoC sequences are tabulated in Table~\ref{tab:euroc} 
which shows that the proposed RI-FLS achieved comparable accuracy to the established method, incremental FLS with a traditional error formulation.
The odometry accuracy could be improved by using the concept of keyframes as in \cite{forster_manifold_2017}.

\begin{table}[]
	\centering
	\begin{tabular}{lllll}
		\hline
		\textbf{\begin{tabular}[c]{@{}l@{}}Mean ATE\\ RMS (m)\end{tabular}}  & \textbf{MH\_01} & \textbf{MH\_05} & \textbf{V1\_02} & \textbf{V2\_02} \\ \hline
		Inc. FLS & 0.88            & \textbf{0.68}  & \textbf{0.28} & 0.24 \\ \hline
		RI-FLS   & \textbf{0.53}   & 0.89           & \textbf{0.28} & 0.29 \\ \hline
		\begin{tabular}[c]{@{}l@{}}RI-FLS with\\ exact Jacobians\end{tabular} 
                         & 0.82           & 1.26           & 0.39         & \textbf{0.23} \\ \hline
	\end{tabular}
	\caption{Absolute translation error RMS averaged over 3 runs on several EuRoC sessions for 
	incremental fixed-lag smoother (FLS), right invariant FLS (RI-FLS, the proposed), RI-FLS with exact IMU factor Jacobians.
	All methods use smart factors to deal with degenerate landmarks. }
	\label{tab:euroc}
\end{table}

\section{Conclusion}
\label{sec:conclusion}

To fix the inconsistent covariances output by traditional FLSs, we introduce the right invariant error formulation into the FLS framework.
We analyze its observability directly with the linearized system, which has much lower analysis complexity than observability matrices.
As a byproduct, we find that landmarks parameterized in a local camera frame and sensor parameters like biases do not affect the estimator consistency.
In the end, we prove that the right invariant error formulation ensures the observability property of a FLS 
without artificially correcting Jacobians like the first estimate Jacobian method.
The proposed right invariant FLS is applied to a monocular visual inertial SLAM problem.
Its consistency is confirmed by simulation, and its practicality is verified with the EuRoC benchmark.

In the future, we will examine the consistency of observable parameters after marginalization,
and look into the properties of the left invariant error formulation.

\section*{Acknowledgments}
We thank the anonymous reviewers for stimulating comments and suggestions.
Jianzhu Huai is partially funded by the National Natural Science Foundation of China (grant number 62003248).
\bibliography{huai}

\begin{thebibliography}{35}
\providecommand{\natexlab}[1]{#1}
\providecommand{\url}[1]{\texttt{#1}}
\providecommand{\urlprefix}{URL }
\expandafter\ifx\csname urlstyle\endcsname\relax
  \providecommand{\doi}[1]{doi:\discretionary{}{}{}#1}\else
  \providecommand{\doi}{doi:\discretionary{}{}{}\begingroup
  \urlstyle{rm}\Url}\fi

\bibitem[{{Bar-Shalom}, Li, and
  Kirubarajan(2004)}]{bar-shalomEstimationApplicationsTracking2004}
{Bar-Shalom}, Y.; Li, X.; and Kirubarajan, T. 2004.
\newblock \emph{Estimation with Applications to Tracking and Navigation: Theory
  Algorithms and Software}.
\newblock {John Wiley \& Sons}.

\bibitem[{Barfoot and Furgale(2014)}]{barfootAssociating2014}
Barfoot, T.~D.; and Furgale, P.~T. 2014.
\newblock Associating uncertainty with three-dimensional poses for use in
  estimation problems.
\newblock \emph{IEEE Transactions on Robotics} 30(3): 679--693.

\bibitem[{Barrau and Bonnabel(2016{\natexlab{a}})}]{barrau_ekf_2016}
Barrau, A.; and Bonnabel, S. 2016{\natexlab{a}}.
\newblock An {{EKF}}-{{SLAM}} algorithm with consistency properties.
\newblock Technical report.
\newblock \urlprefix\url{http://arxiv.org/abs/1510.06263}.

\bibitem[{Barrau and
  Bonnabel(2016{\natexlab{b}})}]{barrauInvariantExtendedKalman2016}
Barrau, A.; and Bonnabel, S. 2016{\natexlab{b}}.
\newblock The invariant extended {{Kalman}} filter as a stable observer.
\newblock \emph{IEEE Transactions on Automatic Control} 62(4): 1797--1812.

\bibitem[{Brossard, Barrau, and
  Bonnabel(2018)}]{brossardExploitingSymmetriesDesign2018}
Brossard, M.; Barrau, A.; and Bonnabel, S. 2018.
\newblock Exploiting Symmetries to Design {{EKFs}} with Consistency Properties
  for Navigation and {{SLAM}}.
\newblock \emph{IEEE Sensors Journal} 19(4): 1572--1579.

\bibitem[{Brossard et~al.(2020)Brossard, Barrau, Chauchat, and
  Bonnabel}]{brossardAssociatingUncertaintyExtended2020}
Brossard, M.; Barrau, A.; Chauchat, P.; and Bonnabel, S. 2020.
\newblock Associating uncertainty to extended poses for on {{Lie}} group IMU
  preintegration with rotating {{Earth}}.
\newblock Technical report.
\newblock \urlprefix\url{http://arxiv.org/abs/2007.14097}.

\bibitem[{Burri et~al.(2016)Burri, Nikolic, Gohl, Schneider, Rehder, Omari,
  Achtelik, and Siegwart}]{burri2016euroc}
Burri, M.; Nikolic, J.; Gohl, P.; Schneider, T.; Rehder, J.; Omari, S.;
  Achtelik, M.~W.; and Siegwart, R. 2016.
\newblock The {{EuRoC}} micro aerial vehicle datasets.
\newblock \emph{The International Journal of Robotics Research} 35(10):
  1157--1163.

\bibitem[{Castellanos et~al.(2007)Castellanos, Martinez-Cantin, Tard{\'o}s, and
  Neira}]{castellanosRobocentricMapJoining2007}
Castellanos, J.; Martinez-Cantin, R.; Tard{\'o}s, J.; and Neira, J. 2007.
\newblock Robocentric map joining: {{Improving}} the consistency of
  {{EKF}}-{{SLAM}}.
\newblock \emph{Robotics and Autonomous Systems} 55(1): 21--29.
\newblock
  \urlprefix\url{https://linkinghub.elsevier.com/retrieve/pii/S0921889006001448}.

\bibitem[{Civera, Davison, and
  Montiel(2008)}]{civeraInverseDepthParametrization2008}
Civera, J.; Davison, A.; and Montiel, J. 2008.
\newblock Inverse depth parametrization for monocular {{SLAM}}.
\newblock \emph{IEEE Transactions on Robotics} 24(5): 932--945.
\newblock \doi{10.1109/TRO.2008.2003276}.

\bibitem[{Costante and
  Mancini(2020)}]{costanteUncertaintyEstimationDatadriven2020}
Costante, G.; and Mancini, M. 2020.
\newblock Uncertainty estimation for data-driven visual odometry.
\newblock \emph{IEEE Transactions on Robotics} 36(6): 1738--1757.
\newblock \doi{10.1109/TRO.2020.3001674}.

\bibitem[{Dellaert(2012)}]{dellaertFactorGraphsGTSAM2012}
Dellaert, F. 2012.
\newblock Factor graphs and {{GTSAM}}: {{A}} hands-on introduction.
\newblock Technical Report GT-RIM-CP\&R-2012-002, {Georgia Institute of
  Technology}, {Atlanta, Georgia, US}.

\bibitem[{{Dong-Si} and Mourikis(2011)}]{dong-siMotionTrackingFixedlag2011}
{Dong-Si}, T.-C.; and Mourikis, A.~I. 2011.
\newblock Motion tracking with fixed-lag smoothing: {{Algorithm}} and
  consistency analysis.
\newblock In \emph{2011 {{IEEE International Conference}} on {{Robotics}} and
  {{Automation (ICRA)}}}, 5655--5662. {Shanghai, China}.

\bibitem[{{Dong-Si} and
  Mourikis(2012)}]{dong-siConsistencyAnalysisSlidingwindow2012}
{Dong-Si}, T.-C.; and Mourikis, A.~I. 2012.
\newblock Consistency analysis for sliding-window visual odometry.
\newblock In \emph{2012 {{IEEE International Conference}} on {{Robotics}} and
  {{Automation (ICRA)}}}, 5202--5209. {Saint Paul, MN, USA}.

\bibitem[{Forster et~al.(2017)Forster, Carlone, Dellaert, and
  Scaramuzza}]{forster_manifold_2017}
Forster, C.; Carlone, L.; Dellaert, F.; and Scaramuzza, D. 2017.
\newblock On-manifold preintegration for real-time visual-inertial odometry.
\newblock \emph{IEEE Transactions on Robotics} 33(1): 1--21.
\newblock \urlprefix\url{https://ieeexplore.ieee.org/document/7557075/}.

\bibitem[{Heo and Park(2018)}]{heoConsistentEKFbasedVisualinertial2018}
Heo, S.; and Park, C.~G. 2018.
\newblock Consistent {{EKF}}-based visual-inertial odometry on matrix {{Lie}}
  group.
\newblock \emph{IEEE Sensors Journal} 18(9): 3780--3788.

\bibitem[{Hermann and
  Krener(1977)}]{hermannNonlinearControllabilityObservability1977}
Hermann, R.; and Krener, A. 1977.
\newblock Nonlinear controllability and observability.
\newblock \emph{IEEE Transactions on Automatic Control} 22(5): 728--740.
\newblock \doi{10.1109/TAC.1977.1101601}.

\bibitem[{Hesch et~al.(2014{\natexlab{a}})Hesch, Kottas, Bowman, and
  Roumeliotis}]{heschCameraIMUbasedLocalizationObservability2014}
Hesch, J.~A.; Kottas, D.~G.; Bowman, S.~L.; and Roumeliotis, S.~I.
  2014{\natexlab{a}}.
\newblock Camera-{{IMU}}-based localization: {{Observability}} analysis and
  consistency improvement.
\newblock \emph{The International Journal of Robotics Research} 33(1):
  182--201.
\newblock \doi{10.1177/0278364913509675}.

\bibitem[{Hesch et~al.(2014{\natexlab{b}})Hesch, Kottas, Bowman, and
  Roumeliotis}]{heschConsistencyAnalysisImprovement2014}
Hesch, J.~A.; Kottas, D.~G.; Bowman, S.~L.; and Roumeliotis, S.~I.
  2014{\natexlab{b}}.
\newblock Consistency analysis and improvement of vision-aided inertial
  navigation.
\newblock \emph{IEEE Transactions on Robotics} 30(1): 158--176.
\newblock \doi{10.1109/TRO.2013.2277549}.

\bibitem[{Huang, Mourikis, and
  Roumeliotis(2010)}]{huangObservabilitybasedRulesDesigning2010}
Huang, G.~P.; Mourikis, A.~I.; and Roumeliotis, S.~I. 2010.
\newblock Observability-based rules for designing consistent {{EKF SLAM}}
  estimators.
\newblock \emph{The International Journal of Robotics Research} 29(5):
  502--528.

\bibitem[{Jekeli(2001)}]{jekeli_inertial_2001}
Jekeli, C. 2001.
\newblock \emph{Inertial {{Navigation Systems}} with {{Geodetic
  Applications}}}.
\newblock {Berlin, Germany}: {de Gruyter}.
\newblock \doi{10.1515/9783110800234}.

\bibitem[{Jones and Soatto(2011)}]{jones2011visual}
Jones, E.~S.; and Soatto, S. 2011.
\newblock Visual-inertial navigation, mapping and localization: A scalable
  real-time causal approach.
\newblock \emph{The International Journal of Robotics Research} 30(4):
  407--430.

\bibitem[{Jung, Heo, and Park(2020)}]{jungObservability2020}
Jung, J.~H.; Heo, S.; and Park, C.~G. 2020.
\newblock Observability analysis of {{IMU}} intrinsic parameters in stereo
  visual\textendash inertial odometry.
\newblock \emph{IEEE Transactions on Instrumentation and Measurement} 69(10):
  7530--7541.
\newblock \doi{10.1109/TIM.2020.2985174}.

\bibitem[{Kaess et~al.(2012)Kaess, Johannsson, Roberts, Ila, Leonard, and
  Dellaert}]{kaess2012isam2}
Kaess, M.; Johannsson, H.; Roberts, R.; Ila, V.; Leonard, J.~J.; and Dellaert,
  F. 2012.
\newblock {{iSAM2}}: {{Incremental}} smoothing and mapping using the {{Bayes}}
  tree.
\newblock \emph{The International Journal of Robotics Research} 31(2):
  216--235.

\bibitem[{Kelly and Sukhatme(2011)}]{kelly2011visual}
Kelly, J.; and Sukhatme, G.~S. 2011.
\newblock Visual-inertial sensor fusion: Localization, mapping and
  sensor-to-sensor self-calibration.
\newblock \emph{The International Journal of Robotics Research} 30(1): 56--79.

\bibitem[{Leutenegger et~al.(2015)Leutenegger, Lynen, Bosse, Siegwart, and
  Furgale}]{leutenegger_keyframe_2015}
Leutenegger, S.; Lynen, S.; Bosse, M.; Siegwart, R.; and Furgale, P. 2015.
\newblock Keyframe-based visual-inertial odometry using nonlinear optimization.
\newblock \emph{The International Journal of Robotics Research} 34(3):
  314--334.
\newblock \urlprefix\url{https://doi.org/10.1177/0278364914554813}.

\bibitem[{Li and Mourikis(2013)}]{li_high_2013}
Li, M.; and Mourikis, A.~I. 2013.
\newblock High-precision, consistent {EKF}-based visual-inertial odometry.
\newblock \emph{The International Journal of Robotics Research} 32(6):
  690--711.
\newblock \urlprefix\url{https://doi.org/10.1177/0278364913481251}.

\bibitem[{Mirzaei and Roumeliotis(2008)}]{mirzaei_kalman_2008}
Mirzaei, F.~M.; and Roumeliotis, S.~I. 2008.
\newblock A {{Kalman}} filter-based algorithm for {{IMU}}-camera calibration:
  {{Observability}} analysis and performance evaluation.
\newblock \emph{IEEE Transactions on Robotics} 24(5): 1143--1156.
\newblock \doi{10.1109/TRO.2008.2004486}.

\bibitem[{Polok et~al.(2015)Polok, Lui, Ila, Drummond, and
  Mahony}]{polokEffect2015}
Polok, L.; Lui, V.; Ila, V.; Drummond, T.; and Mahony, R. 2015.
\newblock The effect of different parameterisations in incremental structure
  from motion.
\newblock In \emph{2015 Australian Conference on {{Robotics}} and
  {{Automation}} ({{ACRA}})}. {Canberra, Australia}.

\bibitem[{Rosinol et~al.(2020)Rosinol, Abate, Chang, and
  Carlone}]{rosinolKimera2020}
Rosinol, A.; Abate, M.; Chang, Y.; and Carlone, L. 2020.
\newblock Kimera: {{An}} open-source library for real-time metric-semantic
  localization and mapping.
\newblock In \emph{2020 {{IEEE International Conference}} on {{Robotics}} and
  {{Automation}} ({{ICRA}})}, 1689--1696. {Paris, France}.
\newblock \urlprefix\url{https://github.com/MIT-SPARK/Kimera}.

\bibitem[{Sol{\`a} et~al.(2012)Sol{\`a}, {Vidal-Calleja}, Civera, and
  Montiel}]{sola2012impact}
Sol{\`a}, J.; {Vidal-Calleja}, T.; Civera, J.; and Montiel, J. M.~M. 2012.
\newblock Impact of landmark parametrization on monocular {{EKF}}-{{SLAM}} with
  points and lines.
\newblock \emph{International Journal of Computer Vision} 97(3): 339--368.
\newblock \doi{10.1007/s11263-011-0492-5}.

\bibitem[{Triggs et~al.(2000)Triggs, McLauchlan, Hartley, and
  Fitzgibbon}]{triggsBundle2000}
Triggs, B.; McLauchlan, P.~F.; Hartley, R.~I.; and Fitzgibbon, A.~W. 2000.
\newblock Bundle adjustment -- {{A}} modern synthesis.
\newblock In Triggs, B.; Zisserman, A.; and Szeliski, R., eds., \emph{Vision
  {{Algorithms}}: {{Theory}} and {{Practice}}}, Lecture {{Notes}} in {{Computer
  Science}}, 298--372. {Berlin, Heidelberg}: {Springer}.
\newblock \doi{10.1007/3-540-44480-7_21}.

\bibitem[{Usenko et~al.(2020)Usenko, Demmel, Schubert, St{\"u}ckler, and
  Cremers}]{usenko_visual_2020}
Usenko, V.; Demmel, N.; Schubert, D.; St{\"u}ckler, J.; and Cremers, D. 2020.
\newblock Visual-inertial mapping with non-linear factor recovery.
\newblock \emph{IEEE Robotics and Automation Letters} 5(2): 422--429.
\newblock \doi{10.1109/LRA.2019.2961227}.

\bibitem[{Yang et~al.(2020)Yang, Geneva, Zuo, and
  Huang}]{yangOnlineImuIntrinsic2020}
Yang, Y.; Geneva, P.; Zuo, X.; and Huang, G. 2020.
\newblock Online {{IMU}} intrinsic calibration: {{Is}} it necessary?
\newblock In \emph{Robotics: {{Science}} and {{Systems}} ({{RSS}})}, 716--725.
  {Corvallis, Oregon}.

\bibitem[{Zhang et~al.(2017)Zhang, Wu, Song, Huang, and
  Dissanayake}]{zhangConvergenceConsistencyAnalysis2017}
Zhang, T.; Wu, K.; Song, J.; Huang, S.; and Dissanayake, G. 2017.
\newblock Convergence and consistency analysis for a 3-{{D Invariant}}-{{EKF
  SLAM}}.
\newblock \emph{IEEE Robotics and Automation Letters} 2(2): 733--740.

\bibitem[{Zhang and Scaramuzza(2018)}]{zhang2018tutorial}
Zhang, Z.; and Scaramuzza, D. 2018.
\newblock A tutorial on quantitative trajectory evaluation for
  visual(-inertial) odometry.
\newblock In \emph{2018 {{IEEE}}/{{RSJ}} International Conference on
  {{Intelligent Robots}} and {{Systems}} ({{IROS}})}, 7244--7251. {Madrid,
  Spain}.
\newblock \doi{10.1109/IROS.2018.8593941}.

\end{thebibliography}

\twocolumn[\centering \section*{\LARGE \bf Supplementary Material \\ \enskip}]
\noindent
In the supplementary material, we formally prove the following assertions for the visual inertial SLAM problem:
\begin{itemize}
\item Sensor parameters and landmarks expressed in local coordinates do not 
affect the unobservable directions.
\item With a traditional error formulation, the unobservable rotational
direction becomes wrongly observable after a marginalization step.
\item With the right invariant error formulation, the unobservable
directions are invariant to the marginalization.
\end{itemize}

\section*{A. Nullspace of the Linearized System}
To prepare for subsequent proofs, this section introduces extra expressions based on the discussion in the main text.

Recall that a transformation of a system, $\mathcal{T}_\xi$, transforms variables from one world frame to another.
It acts on only variables in the world frame and does not 
bother with landmarks expressed in the local camera frame and sensor parameters, \eg, IMU biases.

Prior to any marginalization, when the visual inertial SLAM problem 
goes through a transformation $\mathcal{T}_\xi$ on the unobservable directions,
the residual errors do not change (14), and hence the objective function of the problem, to the extent allowed by noises and first order approximation.
This in turn implies that the linearized system is not affected by a
small change in the state variables induced by the transformation (see (15)).
The net result is that the nullspace $\mathbf{N}_J$ of the coefficient
matrix $\mathbf{J}$ corresponds to unobservable directions of the system.

For the visual inertial SLAM problem, $\mathbf{N}_J$ has four columns
corresponding to the unobservable 1 DOF (degree of freedom) rotation about gravity and 3 DOF absolute translation.
And the unobservable $\mathcal{T}_\xi$ is minimally parameterized by $\boldsymbol{\xi}$
corresponding to the four unobservable directions, \ie,
$\boldsymbol{\xi} = [\delta \phi \quad \delta \mathbf{t}]$
where $\delta \phi$ is a `small' rotation about gravity, and $\delta \mathbf{t}$ a `small' translation of the world frame.

The effect of $\mathcal{T}_\xi$ on a system state variable $\mathbf{x}_i$ is given by
\begin{equation}
\begin{split}
\mathcal{T}_\xi(\mathbf{x}_i) &= (\mathbf{R}(\mathbf{g}\delta\phi)\mathbf{R}_i, \mathbf{R}(\mathbf{g}\delta\phi) \mathbf{v}_i,\\
&\quad\quad \mathbf{R}(\mathbf{g}\delta \phi)\mathbf{p}_i + \delta \mathbf{t}, \mathbf{b}_i) \\
&\stackrel{(23)}{=} (\exp(\mathcal{L}(\mathbf{g} \delta\phi, \mathbf{0}, \delta \mathbf{t}))\boldsymbol{\pi}_i, \mathbf{b}_i)
\end{split}
\end{equation}
where $\mathbf{R}(\mathbf{g}\delta\phi) = \exp((\mathbf{g}\delta \phi)_\times)$.

For system variables up to $t_k$,
$\mathcal{X}_k \stackrel{(2)}{=} (\mathbf{x}_0, \mathbf{x}_1, \dots,  \mathbf{x}_k, \mathbf{f}_1, \mathbf{f}_2, \dots, \mathbf{f}_L)$,
the nullspace of the linearized system, $\mathbf{N}_J$, can be written as 
\begin{equation}
\label{eq:N_J}
\mathbf{N}_J \stackrel{(16)}{=} \begin{bmatrix}
\mathbf{N}_{x_0} \\
\mathbf{N}_{x_1} \\
\vdots \\
\mathbf{N}_{x_k} \\
\mathbf{N}_{f_1} \\
\vdots \\
\mathbf{N}_{f_L} 
\end{bmatrix}
\end{equation}
where the components of $\mathbf{N}_J$ correspond to variables in $\mathcal{X}_k$.
It is worth noting that $\mathbf{N}_J$ is always evaluated at the latest estimate of $\mathcal{X}_k$, 
unless some technique like ``first estimate Jacobians (FEJ)'' \cite{huangObservabilitybasedRulesDesigning2010} is used.
For $\mathbf{x}_i = (\boldsymbol{\pi}_i, \mathbf{b}_i)$, we also write
\begin{equation}
\mathbf{N}_{x_i} = \begin{bmatrix}
\mathbf{N}_{\pi_i} \\
\mathbf{N}_{b_i}
\end{bmatrix}
\end{equation}
As $\mathcal{T}_{\xi}$ does not act on local parameters, by the definition of $\mathbf{N}_J$ (16), we see that
\begin{equation}
\label{eq:N_components}
\begin{split}
\mathbf{N}_{f_i} = \mathbf{0} \quad i \in [1, 2, \dots, L] \\
\mathbf{N}_{b_i} = \mathbf{0} \quad i \in [0, 1, \dots, k],
\end{split}
\end{equation}
and $\mathbf{N}_{\pi_i}$ depends on only $\boldsymbol{\pi}_i$.
Before any marginalization occurs, as the system maintains unobservable directions, the nullspace of $\mathbf{J}$, $\mathbf{N}_J$ has four columns.

\section*{B. Local Parameters Are Irrelevant}
\label{sec:local-params}
This section proves that local state variables including IMU biases,
landmarks expressed in a local camera frame, and camera parameters, do not affect $\mathbf{N}_J$.
This argument uses two assumptions:
\begin{enumerate}
\item The noises do not interfere with the observability analysis.
\item Many derivatives are obtained with the first order approximation.
\end{enumerate}
Assumption (1) is common practice in observability analysis, \eg, \cite{hermannNonlinearControllabilityObservability1977}.
Assumption (2) is even more prevalent, \eg, \cite{barrauInvariantExtendedKalman2016}.

It suffices to prove that for any block row of $\mathbf{J}$, $\mathbf{J}_{(r, :)}$,
whether the nullity $\mathbf{J}_{(r, :)} \mathbf{N}_J = \mathbf{0}$ is satisfied depends on 
only the linearization points of navigation state variables $\boldsymbol{\pi}_i$.
Without loss of generality, we suppose the residual corresponding to $\mathbf{J}_{(r, :)}$ involves only three variables, $\boldsymbol{\pi}_i$, $\mathbf{b}_i$, and $\mathbf{f}_j$.
The nullity can be analyzed in two cases based on whether the residual has been permanently linearized or not in marginalization.
If the residual is not among the linearized terms (see (18)), 
the nullity is satisfied due to the observability property of the nonlinear system (17), \ie.
\begin{equation}
\label{eq:JNnormal}
\mathbf{J}_{(r, :)}(\bar{\boldsymbol{\pi}}_i, \bar{\mathbf{b}}_i, \bar{\mathbf{f}}_j) \mathbf{N}_J(\bar{\boldsymbol{\pi}}_i, \bar{\mathbf{b}}_i, \bar{\mathbf{f}}_j) = \mathbf{0}
\end{equation}
where  $\bar{\boldsymbol{\pi}}_i$, $\bar{\mathbf{b}}_i$, and $\bar{\mathbf{f}}_j$ are the latest estimates of the variables.

Otherwise, $\mathbf{J}_{(r, :)}$ is evaluated at a set of earlier estimates $(\bar{\boldsymbol{\pi}}_i, \bar{\mathbf{b}}_i, \bar{\mathbf{f}}_j)$, 
and $\mathbf{N}_J$ is in general evaluated at the latest estimates $(\bar{\boldsymbol{\pi}}'_i, \bar{\mathbf{b}}'_i, \bar{\mathbf{f}}'_j)$,
the nullity equation becomes
\begin{equation}
\label{eq:JNinit}
\begin{split}
\mathbf{J}_{(r, :)}(\bar{\boldsymbol{\pi}}_i, \bar{\mathbf{b}}_i, \bar{\mathbf{f}}_j) \mathbf{N}_J(\bar{\boldsymbol{\pi}}'_i, \bar{\mathbf{b}}'_i, \bar{\mathbf{f}}'_j) \stackrel{\eqref{eq:N_components}}{=} \\
\mathbf{J}_{(r, :)}(\bar{\boldsymbol{\pi}}_i, \bar{\mathbf{b}}_i, \bar{\mathbf{f}}_j) \mathbf{N}_J(\bar{\boldsymbol{\pi}}'_i)
\end{split}
\end{equation}

With a few optimization iterations, these variables are updated to 
($\bar{\boldsymbol{\pi}}''_i, \bar{\mathbf{b}}''_i, \bar{\mathbf{f}}''_j)$,
and the nullity equation changes to
\begin{equation}
\label{eq:JNiter}
\begin{split}
\mathbf{J}_{(r, :)}(\bar{\boldsymbol{\pi}}_i, \bar{\mathbf{b}}_i, \bar{\mathbf{f}}_j) \mathbf{N}_J(\bar{\boldsymbol{\pi}}''_i, \bar{\mathbf{b}}''_i, \bar{\mathbf{f}}''_j)
\stackrel{\eqref{eq:N_components}}{=} \\
\mathbf{J}_{(r, :)}(\bar{\boldsymbol{\pi}}_i, \bar{\mathbf{b}}_i, \bar{\mathbf{f}}_j) \mathbf{N}_J(\bar{\boldsymbol{\pi}}''_i)
\end{split}
\end{equation}
where the evaluation points for $\mathbf{J}_{(r, :)}$ remain fixed because the residual has been permanently linearized.

From \eqref{eq:JNinit} and \eqref{eq:JNiter}, we see that the nullity condition only depends on the evolving estimates of $\boldsymbol{\pi}_i$. 
Though we may update estimates for $\mathbf{b}_i$ and $\mathbf{f}_j$, they do not help nullifying \eqref{eq:JNinit} or \eqref{eq:JNiter}.
By summarizing the two cases, we conclude that these local parameters do not impact the unobservable directions.
For the FEJ technique, this indicates that
we do not need to use ``first estimates'' for landmarks in a local frame, biases, or camera extrinsic parameters.

\section*{C. Shrunk Nullspace of Traditional Errors}
\label{sec:traditional-errors}
This section proves that nullspace of the linearized system for traditional errors shrinks in the rotation direction after marginalization.
Without loss of generality, we choose the same set of errors as in
\cite{li_high_2013} and \cite{leutenegger_keyframe_2015}.
For other error definitions, \eg, those in \cite{forster_manifold_2017}, the proof goes similarly.
Besides the assumptions in Section \ref{sec:local-params}, we will use the additional assumptions,
\begin{enumerate}
\item The latest state estimates are used to evaluate Jacobians.
\item The component Jacobians for the IMU residual are identities (25).
\end{enumerate}
The former is the best choice for causal estimation.
The latter has been used by \cite{dong-siMotionTrackingFixedlag2011} in proving the consistency of the FLS with the FEJ.

\subsection*{C.1. The Error State and Nullspace}
The traditional error state is defined by
\begin{equation}
\label{eq:traditional-error}
\begin{split}
\delta \mathbf{x}_i &= \boldsymbol{\eta}(\mathbf{x}_i, \bar{\mathbf{x}}_i) \\
&= (\mathcal{L}^{-1}(\log(\mathbf{R}_i \bar{\mathbf{R}}_i^\intercal)), 
\mathbf{v}_i - \bar{\mathbf{v}}_i, \mathbf{p}_i - \bar{\mathbf{p}}_i, \mathbf{b}_i - \bar{\mathbf{b}}_i) \\
\mathbf{x}_i &= \boldsymbol{\eta}^{-1}(\bar{\mathbf{x}}_i, \delta \mathbf{x}_i)
\end{split}
\end{equation}
where $\log(\cdot)$ is the logarithm map at the identity for the special orthogonal group, $SO(3)$.
The inverse Lie operator $\mathcal{L}^{-1}$ converts a skew-symmetric matrix to the corresponding 3D vector.
The key component of the nullspace of $\mathbf{J}$ is
\begin{equation}
\mathbf{N}_{\pi_i} = \frac{\partial \boldsymbol{\eta}(\mathcal{T}_\xi(\boldsymbol{\pi}_i), \bar{\boldsymbol{\pi}}_i)}{\partial \boldsymbol{\xi}} \bigg\rvert_{\bar{\boldsymbol{\pi}}_i} \\
= \begin{bmatrix}
\mathbf{g} & \mathbf{0} \\
-(\bar{\mathbf{v}}_i)_\times\mathbf{g} & \mathbf{0} \\
-(\bar{\mathbf{p}}_i)_\times\mathbf{g} & \mathbf{I}_3
\end{bmatrix}
\end{equation}

\subsection*{C.2. Residual Errors and Jacobians}
For the IMU residual error, $\mathbf{r}_x(\mathbf{x}_i, \mathbf{x}_{i|i-1}) = \boldsymbol{\eta}(\mathbf{x}_i, \mathbf{x}_{i|i-1})$, 
its Jacobians components $\mathbf{A}_i$ and $\mathbf{A}_{i|i-1}$ (12) are given by
\begin{equation}
\begin{split}
\mathbf{A}_i = \begin{bmatrix}
\mathbf{J}_l^{-1}(\mathcal{L}^{-1}(\log(\mathbf{R}_i \bar{\mathbf{R}}_i^\intercal))) & & & \\
& \mathbf{I}_3 & & \\
& & \mathbf{I}_3 & \\
& & & \mathbf{I}_6
\end{bmatrix} \\
\mathbf{A}_{i|i-1} = -\begin{bmatrix}
\mathbf{J}_l^{-1}(-\mathcal{L}^{-1}(\log(\mathbf{R}_i \bar{\mathbf{R}}_i^\intercal))) & & & \\
& \mathbf{I}_3 & & \\
& & \mathbf{I}_3 & \\
& & & \mathbf{I}_6
\end{bmatrix},
\end{split}
\end{equation}
where $\mathbf{J}_l(\cdot)$ is the left Jacobian of $SO(3)$ \cite{barfootAssociating2014}.
The transition matrix $\boldsymbol{\Phi}_{i|i-1}$ is given by
\begin{equation}
\label{eq:transition_mat}
\begin{split}
\boldsymbol{\Phi}_{i|i-1} =
\begin{bmatrix}
\boldsymbol{\Phi}_\pi & \boldsymbol{\Phi}_{\pi, b} \\
\mathbf{0} & \mathbf{I}
\end{bmatrix}
\end{split}	
\end{equation}
$\boldsymbol{\Phi}_{\pi, i|i-1}$ can be written out as
\begin{align}
\boldsymbol{\Phi}_{\pi, i|i-1} &= \begin{bmatrix}
\mathbf{I}_3 & \mathbf{0}  & \mathbf{0} \\
\boldsymbol{\Phi}_{vq} & \mathbf{I}_3 & \mathbf{0} \\
\boldsymbol{\Phi}_{pq}&  \mathbf{I}(t_{i} - t_{i-1}) & \mathbf{I}_3
\end{bmatrix} \\
\begin{split}
\boldsymbol{\Phi}_{vq, i|i-1} &= -[\mathbf{v}_{i, s_1} - \mathbf{v}_{i-1, s_1} - \mathbf{g}(t_{i} - t_{i-1})]_\times
\end{split} \\
\begin{split}
\boldsymbol{\Phi}_{pq,i|i-1} &= -[\mathbf{p}_{i, s_1} - \mathbf{p}_{i-1, s_1} - \\
& \quad \mathbf{v}_{i-1, s_1}(t_i - t_{i-1}) - \frac{1}{2}\mathbf{g}(t_i - t_{i-1})^2]_\times
\end{split}
\end{align}
where the subscript $s_1$ identifies the optimization step.
We do not write out $\boldsymbol{\Phi}_{\pi, b}$ because it does not affect the nullspace dimension (Section \ref{sec:local-params}).
Derivation for $\boldsymbol{\Phi}_{i|i-1}$ can be found in \cite{li_high_2013}.

For the camera reprojection residual (11), when the observing camera frame is the anchor frame, it is
trivial to get the Jacobians relative to $\boldsymbol{\pi}_i$ and $\boldsymbol{\pi}_a$, \ie, 
$\mathbf{J}_{\pi_i, l} = \mathbf{0}$ and $\mathbf{J}_{\pi_a, l} = \mathbf{0}$,
because the observation does not depend on $\boldsymbol{\pi}_i$ or $\boldsymbol{\pi}_a$.
We do not write out $\mathbf{J}_{f_l}$ because it does not affect the nullspace as discussed in Section \ref{sec:local-params}.
For the general case, the observation Jacobians $\mathbf{J}_{\pi_i, l}$ and $\mathbf{J}_{\pi_a, l}$
for landmark $\mathbf{f}_l$ observed in frame $i$ with parameters anchored at frame $a$ are
\begin{equation}
\begin{split}
\mathbf{J}_{\pi_i, l} &= \mathbf{J}_h \mathbf{T}_{BC}^{-1} 
\frac{\partial \mathbf{f}_l^{B_i}}{\partial (\delta \boldsymbol{\theta}_i, \delta \mathbf{v}_i, \delta \mathbf{p}_i)} \\
\frac{\partial \mathbf{f}_l^{B_i}}{\partial (\delta \boldsymbol{\theta}_i, \delta \mathbf{v}_i, \delta \mathbf{p}_i)} &=
\begin{bmatrix}
\mathbf{R}_i^\intercal (\mathbf{f}_{l, 1:3}^{W} - \mathbf{p}_i \rho)_\times & \mathbf{0} & -\mathbf{R}_i^\intercal\rho \\
\mathbf{0}^\intercal & \mathbf{0}^\intercal & \mathbf{0}^\intercal
\end{bmatrix}\\
\mathbf{J}_{\pi_a, l} &=  \mathbf{J}_h \mathbf{T}_{BC}^{-1} \mathbf{T}_{WB_i}^{-1} 
\frac{\partial \mathbf{f}_l^{W}}{\partial (\delta \boldsymbol{\theta}_a, \delta \mathbf{v}_a, \delta \mathbf{p}_a)}\\
\frac{\partial \mathbf{f}_l^{W}}{\partial (\delta \boldsymbol{\theta}_a, \delta \mathbf{v}_a, \delta \mathbf{p}_a)} &= \begin{bmatrix}
-(\mathbf{R}_a \mathbf{f}_{l, 1:3}^{B_a})_\times & \mathbf{0} & \rho\mathbf{I}_3 \\
\mathbf{0}^\intercal & \mathbf{0}^\intercal & \mathbf{0}^\intercal
\end{bmatrix}
\end{split}
\end{equation}
where the point $\mathbf{f}_l$ is expressed by homogeneous coordinates in the coordinate frame signified by its superscript, for instance, 
$\mathbf{f}_l^{W} = \mathbf{T}_{WBa} \mathbf{T}_{BC} \mathbf{f}_l$,
and the subscript 1:3 means taking the first 3 elements of $\mathbf{f}_l$, 
and the projection Jacobian component $\mathbf{J}_h$ is 
\begin{equation}
\label{eq:Jh}
\mathbf{J}_h = \frac{\partial\mathbf{h}(\mathbf{T}_{BC}^{-1}\mathbf{T}_{WBi}^{-1}
\mathbf{T}_{WBa} \mathbf{T}_{BC} \mathbf{f}_l)}{\partial \mathbf{f}_l^{C_i}}.
\end{equation}
Noting that the reprojection residual Jacobians relative to IMU biases, $\mathbf{J}_{b_i, l}$ and $\mathbf{J}_{b_a, l}$, are zero,
the reprojection Jacobians in the main text are given by
\begin{equation}
\label{eq:Jxil}
\mathbf{J}_{x_i, l} = [\mathbf{J}_{\pi_i, l}\enskip \mathbf{0}]\quad
\mathbf{J}_{x_a, l} = [\mathbf{J}_{\pi_a, l}\enskip \mathbf{0}]
\end{equation}

\subsection*{C.3. Nullspace and Marginalization}
It can be shown that with the approximation (25), the below equations hold before any marginalization occurs,
\begin{equation}
\label{eq:null-condition}
\begin{split}
-\boldsymbol{\Phi}_{\pi, i|i-1} \mathbf{N}_{\pi_{i-1}}  + \mathbf{N}_{\pi_{i}} &=
\begin{bmatrix}
\mathbf{0}_{9\times 1} \enskip \mathbf{0}_{9\times 3}
\end{bmatrix} \\
\mathbf{J}_{\pi_i, l} \mathbf{N}_{\pi_i} + \mathbf{J}_{\pi_a, l} \mathbf{N}_{\pi_a} &= 
\begin{bmatrix}
\mathbf{0}_{2\times 1} \enskip \mathbf{0}_{2\times 3}
\end{bmatrix},
\end{split}
\end{equation}
for any IMU residual and any reprojection residual.

After marginalization at $t_m$ (18), the linearization points and Jacobians for the linearized
residuals, $\mathbf{r}_x(\mathbf{x}_{i},  \mathbf{x}_{i|i-1})$, $i \in [0, 1, \dots, m]$,
and $\mathbf{r}_{il}, (i, l)\in\mathcal{M}$, are locked.
Let's consider variables that are involved in both the linearized residuals and the remaining nonlinear residuals, \eg, $\boldsymbol{\pi}_{m}$.
Recall that an optimization iteration usually involves three phases, linearization of the objective function, solving the linear system, and finally updating variables.
In subsequent optimization steps, these variables will get
updated, and the nullspace matrix $\mathbf{N}_J$ evaluated at the updated values will be incompatible to
the fixed Jacobians of the linearized factors, \ie, the nullspace condition \eqref{eq:null-condition} does not hold any more.

The situation is better explained with two such variables, $\boldsymbol{\pi}_m$ and $\boldsymbol{\pi}_a$.
$\boldsymbol{\pi}_m$ is involved in the IMU residual $\mathbf{r}_x(\mathbf{x}_{m}, \mathbf{x}_{m|m-1})$
which is to be linearized in marginalization at $t_m$.
$\boldsymbol{\pi}_a$ is involved in a reprojection residual
$\mathbf{r}_{jl}(\mathbf{x}_j, \mathbf{x}_a, \mathbf{f}_l)$ that is to be linearized at $t_m$ too.
The reason for its linearization is that either the landmark 
$\mathbf{f}_l$ or the pose of the observing frame $\mathbf{x}_j$ is earlier in time than $t_m$.
Let's denote the last optimization step before marginalization by $s_1$,
and an optimization step after marginalization by $s_2$.

At step $s_1$, the nullspace equations in \eqref{eq:null-condition} for
$\mathbf{r}_x(\mathbf{x}_{m}, \mathbf{x}_{m|m-1})$ and $\mathbf{r}_{jl}(\mathbf{x}_j, \mathbf{x}_a, \mathbf{f}_l)$ are
\begin{equation}
\begin{split}
\label{eq:null-condition-s1}
-\boldsymbol{\Phi}_{\pi, m|m-1}(\bar{\boldsymbol{\pi}}_{m, s_1}, 
\bar{\boldsymbol{\pi}}_{m-1, s_1}) \mathbf{N}_{\pi_{m-1}}(\bar{\boldsymbol{\pi}}_{m-1, s_1})  + \\
\mathbf{N}_{\pi_{m}}(\bar{\boldsymbol{\pi}}_{m, s_1})=
\begin{bmatrix}
\mathbf{0}_{9\times 1} \enskip \mathbf{0}_{9\times 3}
\end{bmatrix} \\
\mathbf{J}_{\pi_j, l}(\bar{\boldsymbol{\pi}}_{j, s_1}) \mathbf{N}_{\pi_j}(\bar{\boldsymbol{\pi}}_{j, s_1}) + \mathbf{J}_{\pi_a, l}(\bar{\boldsymbol{\pi}}_{a, s_1}) \mathbf{N}_{\pi_a}(\bar{\boldsymbol{\pi}}_{a, s_1}) \\
= \begin{bmatrix}
\mathbf{0}_{2\times 1} \enskip \mathbf{0}_{2\times 3}
\end{bmatrix},
\end{split}
\end{equation}
where the subscript $s_1$ identifies an estimate at step $s_1$.

After the marginalization step, variables earlier than $t_m$, $\boldsymbol{\pi}_j$ and $\boldsymbol{\pi}_{m-1}$,
will no longer be updated.
But variables $\boldsymbol{\pi}_{m}$ and $\boldsymbol{\pi}_{a}$ are still in the optimization problem, and will be updated in subsequent steps.
Since the two residuals in \eqref{eq:null-condition-s1} are permanently linearized,
their Jacobians, $\boldsymbol{\Phi}_{\pi, m|m-1}(\bar{\boldsymbol{\pi}}_{m, s_1}, 
\bar{\boldsymbol{\pi}}_{m-1, s_1})$, $\mathbf{J}_{\pi_j, l}(\bar{\boldsymbol{\pi}}_{j, s_1})$, and $\mathbf{J}_{\pi_a, l}(\bar{\boldsymbol{\pi}}_{a, s_1})$, will be unchangeable.

At step $s_2$, the nullspace matrix blocks $\mathbf{N}_{\pi_m}$ and $\mathbf{N}_{\pi_a}$ are evaluated at the latest estimates of $\boldsymbol{\pi}_{m}$ and $\boldsymbol{\pi}_{a}$, 
\ie, $\bar{\boldsymbol{\pi}}_{m,s_2}$ and $\bar{\boldsymbol{\pi}}_{a,s_2}$,
and blocks of the coefficient matrix $\mathbf{J}$ (13) for the remaining residuals are also evaluated at these estimates.
For instance, a reprojection residual involving $\boldsymbol{\pi}_m$ and a landmark $\mathbf{f}_p$, that has not been linearized,
has a nullspace equation like
\begin{equation}
\begin{split}
\label{eq:null-condition-reproj}
\mathbf{J}_{\pi_m, p}(\bar{\boldsymbol{\pi}}_{m, s_2}) \mathbf{N}_{\pi_m}(\bar{\boldsymbol{\pi}}_{m, s_2}) + \\ \mathbf{J}_{\pi_a, p}(\bar{\boldsymbol{\pi}}_{a, s_2}) \mathbf{N}_{\pi_a}(\bar{\boldsymbol{\pi}}_{a, s_2})
= \begin{bmatrix}
\mathbf{0}_{2\times 1} \enskip \mathbf{0}_{2\times 3}
\end{bmatrix}.
\end{split}
\end{equation}
In contrast, the nullspace matrix blocks $\mathbf{N}_{\pi_j}$ and $\mathbf{N}_{\pi_{m-1}}$ are evaluated at estimates at $s_1$,
\ie, $\bar{\boldsymbol{\pi}}_{j,s_1}$ and $\bar{\boldsymbol{\pi}}_{m-1,s_1}$,
because $\boldsymbol{\pi}_{j}$ and $\boldsymbol{\pi}_{m-1}$ are no longer updated since $s_1$.

In summary, at step $s_2$, the nullspace equations for $\mathbf{r}_x(\mathbf{x}_{m}, \mathbf{x}_{m|m-1})$ and $\mathbf{r}_{jl}(\mathbf{x}_j, \mathbf{x}_a, \mathbf{f}_l)$ are
\begin{equation}
\label{eq:null-condition-failed}
\begin{split}
-\boldsymbol{\Phi}_{\pi, m|m-1}(\bar{\boldsymbol{\pi}}_{m, s_1}, 
\bar{\boldsymbol{\pi}}_{m-1, s_1}) \mathbf{N}_{\pi_{m-1}}(\bar{\boldsymbol{\pi}}_{m-1, s_1})  + \\
\mathbf{N}_{\pi_{m}}(\bar{\boldsymbol{\pi}}_{m, s_2})=
\begin{bmatrix}
\mathbf{d}_1 \enskip \mathbf{0}_{9\times 3}
\end{bmatrix} \\
\mathbf{J}_{\pi_j, l}(\bar{\boldsymbol{\pi}}_{j, s_1}) \mathbf{N}_{\pi_j}(\bar{\boldsymbol{\pi}}_{j, s_1}) + \mathbf{J}_{\pi_a, l}(\bar{\boldsymbol{\pi}}_{a, s_1}) \mathbf{N}_{\pi_a}(\bar{\boldsymbol{\pi}}_{a, s_2}) \\
= \begin{bmatrix}
\mathbf{d}_2 \enskip \mathbf{0}_{2\times 3}
\end{bmatrix},
\end{split}
\end{equation}
where $\mathbf{d}_1$ and $\mathbf{d}_2$ are nonzero vectors due to the combination of different linearization points at $s_1$ and $s_2$.
That is, the nullspace dimension corresponding to the rotation about gravity disappears.

The FEJ technique \cite{huangObservabilitybasedRulesDesigning2010} ensures the dimension of the nullspace of $\mathbf{J}$ (13) 
by using the so-called ``first estimates'' to evaluate blocks of $\mathbf{J}$ and $\mathbf{N}_J$.
The first estimate of a variable is its latest estimate if it is not related to any permanently linearized residual.
Otherwise, its first estimate is set to its latest estimate when
the first residual involving the variable is permanently linearized, and will remain fixed since then.
This way, all blocks in $\mathbf{J}$ and $\mathbf{N}_J$ are evaluated at the same points, \ie, first estimates,
therefore the nullspace shrinkage problem is avoided.
If the FEJ method is applied to the above scenario, the nullspace equations at step $s_2$ for the two considered residuals 
will be the same as \eqref{eq:null-condition-s1} since 
the first estimates $\bar{\boldsymbol{\pi}}_{m, s_1}$ and $\bar{\boldsymbol{\pi}}_{a, s_1}$
are used instead of the latest estimates
$\bar{\boldsymbol{\pi}}_{m, s_2}$ and $\bar{\boldsymbol{\pi}}_{a, s_2}$.
But for a residual that is not linearized, \eg, the one in \eqref{eq:null-condition-reproj}, the nullspace condition will become
\begin{equation}
\label{eq:null-condition-reproj-fej}
\begin{split}
\mathbf{J}_{\pi_m, p}(\bar{\boldsymbol{\pi}}_{m, s_1}) \mathbf{N}_{\pi_m}(\bar{\boldsymbol{\pi}}_{m, s_1}) + \\ \mathbf{J}_{\pi_a, p}(\bar{\boldsymbol{\pi}}_{a, s_1}) \mathbf{N}_{\pi_a}(\bar{\boldsymbol{\pi}}_{a, s_1})
= \begin{bmatrix}
\mathbf{0}_{2\times 1} \enskip \mathbf{0}_{2\times 3}
\end{bmatrix},
\end{split}
\end{equation}
where again the first estimates $\bar{\boldsymbol{\pi}}_{m, s_1}$ and $\bar{\boldsymbol{\pi}}_{a, s_1}$ are used to compute Jacobians. 
By comparing \eqref{eq:null-condition-reproj} and \eqref{eq:null-condition-reproj-fej}, 
we see that the FEJ may hurt accuracy since less accurate values are used for computing Jacobians.

\section*{D. Invariant Nullspace of Right Invariant Errors}
This section shows that the linearized system with right invariant errors
maintains its nullspace dimension after marginalization.

The assumptions are similar to those in Section \ref{sec:traditional-errors}.

\subsection*{D.1. The Error State and Nullspace}
The error definition is given in (20), \ie,
\begin{equation}
\label{eq:invariant-error}
\begin{split}
\delta \mathbf{x}_i &= \boldsymbol{\eta}(\mathbf{x}_i, \bar{\mathbf{x}}_i) \\
&= (\mathcal{L}^{-1}(\log(\boldsymbol{\pi}_i \bar{\boldsymbol{\pi}}_i^\intercal)), \mathbf{b}_i - \bar{\mathbf{b}}_i) \\
\mathbf{x}_i &= \boldsymbol{\eta}^{-1}(\bar{\mathbf{x}}_i, \delta \mathbf{x}_i)
\end{split}
\end{equation}
where $\log(\cdot)$ is the logarithm map for the Lie group $SE_2(3)$ \cite{barrauInvariantExtendedKalman2016}.

The key component of the nullspace of $\mathbf{J}$ is
\begin{equation}
\label{eq:ri_Npi}
\mathbf{N}_{\pi_i} = \frac{\partial \boldsymbol{\eta}(\mathcal{T}_\xi(\boldsymbol{\pi}_i), \bar{\boldsymbol{\pi}}_i)}{\partial \boldsymbol{\xi}} \bigg\rvert_{\bar{\boldsymbol{\pi}}_i} \\
= \begin{bmatrix}
\mathbf{g} & \mathbf{0} \\
\mathbf{0} & \mathbf{0} \\
\mathbf{0} & \mathbf{I}_3
\end{bmatrix}
\end{equation}
As a result, $\mathbf{N}_J$ \eqref{eq:N_J} is independent of the linearization points of the system variables.

\subsection*{D.2. Residual Errors and Jacobians}
The IMU residual error (12) is defined to be
\begin{equation}
\mathbf{r}_x(\mathbf{x}_i, \mathbf{x}_{i|i-1}) =
\boldsymbol{\eta}(\mathbf{x}_i, \mathbf{x}_{i|i-1}).
\end{equation}
The discrete system transition matrix $\boldsymbol{\Phi}_{i|i-1}$ is given by (24).
The Jacobian components $\mathbf{A}_i$ and $\mathbf{A}_{i|i-1}$ are given by
\begin{equation}
\label{eq:A_i}
\begin{split}
\mathbf{A}_i = \begin{bmatrix}
\mathbf{J}_l^{-1}(\mathcal{L}^{-1}(\log(\boldsymbol{\pi}_i \bar{\boldsymbol{\pi}}_i^\intercal))) & \\
& \mathbf{I}_6
\end{bmatrix} \\
\mathbf{A}_{i|i-1} = -\begin{bmatrix}
\mathbf{J}_l^{-1}(-\mathcal{L}^{-1}(\log(\boldsymbol{\pi}_i \bar{\boldsymbol{\pi}}_i^\intercal))) & \\ & \mathbf{I}_6
\end{bmatrix}
\end{split}
\end{equation}
where $\mathbf{J}_l(\cdot)$ is the left Jacobian of $SE_2(3)$.
Its closed form expression can be derived from \cite{barfootAssociating2014}.

For the camera observation (11), when the observing frame is the anchor frame,
the observation Jacobians are trivially zero, \ie, 
$\mathbf{J}_{\pi_i, l} = \mathbf{0}$ and $\mathbf{J}_{\pi_a, l} = \mathbf{0}$.
We do not write out $\mathbf{J}_{f_l}$ because it does not affect the nullspace.
For the general case, the observation Jacobians are
\begin{equation}
\begin{split}
\mathbf{J}_{\pi_i, l} &= \mathbf{J}_h \mathbf{T}_{BC}^{-1} 
\frac{\partial \mathbf{f}_l^{B_i}}{\partial (\delta \boldsymbol{\theta}_i, \delta \mathbf{v}_i, \delta \mathbf{p}_i)} \\
\frac{\partial \mathbf{f}_l^{B_i}}{\partial (\delta \boldsymbol{\theta}_i, \delta \mathbf{v}_i, \delta \mathbf{p}_i)} &=
\begin{bmatrix}
\mathbf{R}_i^\intercal (\mathbf{f}_{l, 1:3}^{W})_\times & \mathbf{0} & -\mathbf{R}_i^\intercal\rho \\
\mathbf{0}^\intercal & \mathbf{0}^\intercal & \mathbf{0}^\intercal
\end{bmatrix}\\
\mathbf{J}_{\pi_a, l} &=  \mathbf{J}_h \mathbf{T}_{BC}^{-1} \mathbf{T}_{WB_i}^{-1} 
\frac{\partial \mathbf{f}_l^{W}}{\partial (\delta \boldsymbol{\theta}_a, \delta \mathbf{v}_a, \delta \mathbf{p}_a)}\\
\frac{\partial \mathbf{f}_l^{W}}{\partial (\delta \boldsymbol{\theta}_a, \delta \mathbf{v}_a, \delta \mathbf{p}_a)} &= \begin{bmatrix}
-(\mathbf{f}_{l, 1:3}^{W})_\times & \mathbf{0} & \rho\mathbf{I}_3 \\
\mathbf{0}^\intercal & \mathbf{0}^\intercal & \mathbf{0}^\intercal
\end{bmatrix},
\end{split}
\end{equation}
where $\mathbf{J}_h$ is given in \eqref{eq:Jh}. 
The reprojection Jacobians blocks in the main text
$\mathbf{J}_{x_i, l}$ and $\mathbf{J}_{x_a, l}$
can be computed with \eqref{eq:Jxil} as in the traditional error case.

\subsection*{D.3. Nullspace and Marginalization}
With the approximation that Jacobian components in \eqref{eq:A_i} are identities, it is
straightforward to verify that after marginalization at $t_m$,
the linearized residuals involving variables that appear in both the linearized residuals and the nonlinear residuals,
still satisfy the nullspace condition.
For the two linearized residuals that cause inconsistency with traditional errors \eqref{eq:null-condition-failed} at step $s_2$,
we have
\begin{equation}
\begin{split}
\label{eq:null-condition-ri}
-\boldsymbol{\Phi}_{\pi, m|m-1}(\bar{\boldsymbol{\pi}}_{m, s_1}, 
\bar{\boldsymbol{\pi}}_{m-1, s_1}) \mathbf{N}_{\pi_{m-1}}(\bar{\boldsymbol{\pi}}_{m-1, s_1})  + \\
\mathbf{N}_{\pi_{m}}(\bar{\boldsymbol{\pi}}_{m, s_2})=
\begin{bmatrix}
\mathbf{0}_{9\times 1} \enskip \mathbf{0}_{9\times 3}
\end{bmatrix} \\
\mathbf{J}_{\pi_j, l}(\bar{\boldsymbol{\pi}}_{j, s_1}) \mathbf{N}_{\pi_j}(\bar{\boldsymbol{\pi}}_{j, s_1}) + \mathbf{J}_{\pi_a, l}(\bar{\boldsymbol{\pi}}_{a, s_1}) \mathbf{N}_{\pi_a}(\bar{\boldsymbol{\pi}}_{a, s_2}) \\
= \begin{bmatrix}
\mathbf{0}_{2\times 1} \enskip \mathbf{0}_{2\times 3}
\end{bmatrix}
\end{split}
\end{equation}
Comparing \eqref{eq:null-condition-ri} with the counterparts for traditional errors, \eqref{eq:null-condition-failed}, we see that the crux for equality in \eqref{eq:null-condition-ri} is that
\begin{equation}
\begin{split}
\mathbf{N}_{\pi_{m}}(\bar{\boldsymbol{\pi}}_{m, s_2}) &\stackrel{\eqref{eq:ri_Npi}}{=} \mathbf{N}_{\pi_{m}}(\bar{\boldsymbol{\pi}}_{m, s_1}) \\
\mathbf{N}_{\pi_a}(\bar{\boldsymbol{\pi}}_{a, s_2}) &\stackrel{\eqref{eq:ri_Npi}}{=} \mathbf{N}_{\pi_a}(\bar{\boldsymbol{\pi}}_{a, s_1}).
\end{split}
\end{equation}
In summary, with right invariant errors, the marginalization step does not introduce spurious
information along the unobservable rotation about gravity.

\subsection*{D.4. Generalization to Multiple Marginalization Steps}
Our analysis only considers one marginalization step, but it is straightforward to extend to multiple steps with the partly linearized objective function (18).
For instance, one marginalization occurs at $t_m$, and the next at $t_n$, then the linearized terms simply expand to include residual errors occurring no later than $t_n$.

\end{document}